\def\confName{CVPR}
\def\confYear{2024}

\def\authorBlock{
	Zhenghao Pan$^{1}\thanks{Equal contribution} $ \qquad Haijin Zeng$^{2}\footnotemark[1]$ \qquad Jiezhang Cao$^{3}$ \qquad Kai Zhang$^{3}$ \qquad Yongyong Chen$^{1}$ \\
	$^{1}$Harbin Institute of Technology \qquad $^{2}$IMEC-UGent \qquad $^{3}$ETH Zurich\\}

\newif\ifreview \newcommand{\review}{\reviewtrue}
\newif\ifarxiv 
\newif\ifcamera 
\newif\ifrebuttal 

\review 

\pdfoutput=1
\documentclass[10pt,twocolumn,letterpaper]{article}
\ifreview \usepackage[final]{cvpr} \fi
\ifarxiv \usepackage[pagenumbers]{cvpr} \fi
\ifrebuttal \usepackage[rebuttal]{cvpr} \fi
\ifcamera \usepackage{cvpr} \fi

\usepackage{graphicx}	
\usepackage{amsmath}
\usepackage{algorithm}
\usepackage{algorithmic}
\usepackage{amssymb}	
\usepackage{booktabs}
\usepackage{times}
\usepackage{microtype}
\usepackage{epsfig}
\usepackage[table,xcdraw,dvipsnames]{xcolor}
\usepackage{multirow, bigstrut}
\usepackage{caption}
\usepackage{float}
\usepackage{placeins}
\usepackage{amsmath}
\usepackage{color, colortbl}
\usepackage{stfloats}
\usepackage{enumitem}
\usepackage{tabularx}
\usepackage{xstring}
\usepackage{multirow}
\usepackage{xspace}
\usepackage{url}
\usepackage{subcaption}
\usepackage{xcolor}
\usepackage{overpic}
\usepackage[hang,flushmargin]{footmisc}
\definecolor{color3}{gray}{0.95}
\definecolor{lightblue}{HTML}{1E90FF} 

\definecolor{lightlightblue}{HTML}{B0E0E6} 
\definecolor{lightred}{HTML}{B22222} 
\definecolor{gree}{HTML}{A8E0B7} 
\definecolor{rouse}{rgb}{0.981,0.961,0.941}
\definecolor{lightgreen}{rgb}{0.9, 0.99, 0.9}
\definecolor{light-yellow}{rgb}{1,1,0.93}
\definecolor{lightgray5}{gray}{0.98}
\definecolor{lightgray6}{gray}{0.96}
\usepackage[pagebackref,breaklinks,colorlinks]{hyperref}

\ifcamera \usepackage[accsupp]{axessibility} \fi

% ...

\newcommand{\h}{0}

\newcommand{\wa}{0.15}
\newlength \g

\usepackage{adjustbox}

\ifarxiv  \fi

\newcommand{\R}[1]{{%
    \textbf{%
        \ifstrequal{#1}{1}{\textcolor{red}{R#1}}{%
        \ifstrequal{#1}{2}{\textcolor{blue}{R#1}}{%
        \ifstrequal{#1}{3}{\textcolor{magenta}{R#1}}{%
        \ifstrequal{#1}{4}{\textcolor{teal}{R#1}}{%
                           \textcolor{cyan}{R#1}%
        }}}}%
    }%
}}

\usepackage{xr-hyper}

\makeatletter
\newcommand*{\addFileDependency}[1]{
  \typeout{(#1)}
  \@addtofilelist{#1}
  \IfFileExists{#1}{}{\typeout{No file #1.}}
}

\makeatother

\begin{document}
\title{DiffSCI: Zero-Shot Snapshot Compressive Imaging via Iterative Spectral Diffusion Model}
\author{\authorBlock}
\maketitle

\begin{abstract}
This paper endeavors to advance the precision of snapshot compressive imaging (SCI) reconstruction for multispectral image (MSI). To achieve this, we integrate the advantageous attributes of established SCI techniques and an image generative model, propose a novel structured zero-shot diffusion model, dubbed DiffSCI. DiffSCI leverages the structural insights from the deep prior and optimization-based methodologies, complemented by the generative capabilities offered by the contemporary denoising diffusion model. Specifically, firstly, we employ a pre-trained diffusion model, which has been trained on a substantial corpus of RGB images, as the generative denoiser within the Plug-and-Play framework for the first time. This integration allows for the successful completion of SCI reconstruction, especially in the case that current methods struggle to address effectively. Secondly, we systematically account for spectral band correlations and introduce a robust methodology to mitigate wavelength mismatch, thus enabling seamless adaptation of the RGB diffusion model to MSIs.
Thirdly, an accelerated algorithm is implemented to expedite the resolution of the data subproblem. This augmentation not only accelerates the convergence rate but also elevates the quality of the reconstruction process.
We present extensive testing to show that DiffSCI exhibits discernible performance enhancements over prevailing self-supervised and zero-shot approaches, surpassing even supervised transformer counterparts across both simulated and real datasets. Our code will be available.

\end{abstract}

\vspace{-4mm}
\section{Introduction}
\label{sec:intro}
\vspace{-2mm}
Contrary to conventional RGB images, multispectral images (MSIs) incorporate an expanded array of spectral bands, enabling the retention of more comprehensive and detailed information. Therefore, MSIs are widely applied in remote sensing~\cite{rs_1,rs_2,9927348,rs_3,rs_4,10128755}, medical imaging~\cite{mi_1,mi_2,mi_3}, environmental monitoring~\cite{thenkabail2014hyperspectral}, etc. Owing to the advancement of snapshot compressive imaging (SCI) systems~\cite{sci_1,sci_2,sci_3,sci_5,sci_6,yuan2015compressive,ma2021led}, it has become feasible to acquire two-dimensional measurements of MSIs. The decoding stage of the SCI system aims to reconstruct the three-dimensional MSIs from its degraded two-dimensional measurement.

\begin{figure}
	\centering
	\renewcommand{\h}{0.105}
	\renewcommand{\wa}{0.12}
	\newcommand{\wb}{0.16}
	\renewcommand{\g}{-0.7mm}
	\renewcommand{\tabcolsep}{1.8pt}
	\renewcommand{\arraystretch}{1}
	 \resizebox{1\linewidth}{!} {\includegraphics[width=0.47\linewidth, angle =270]{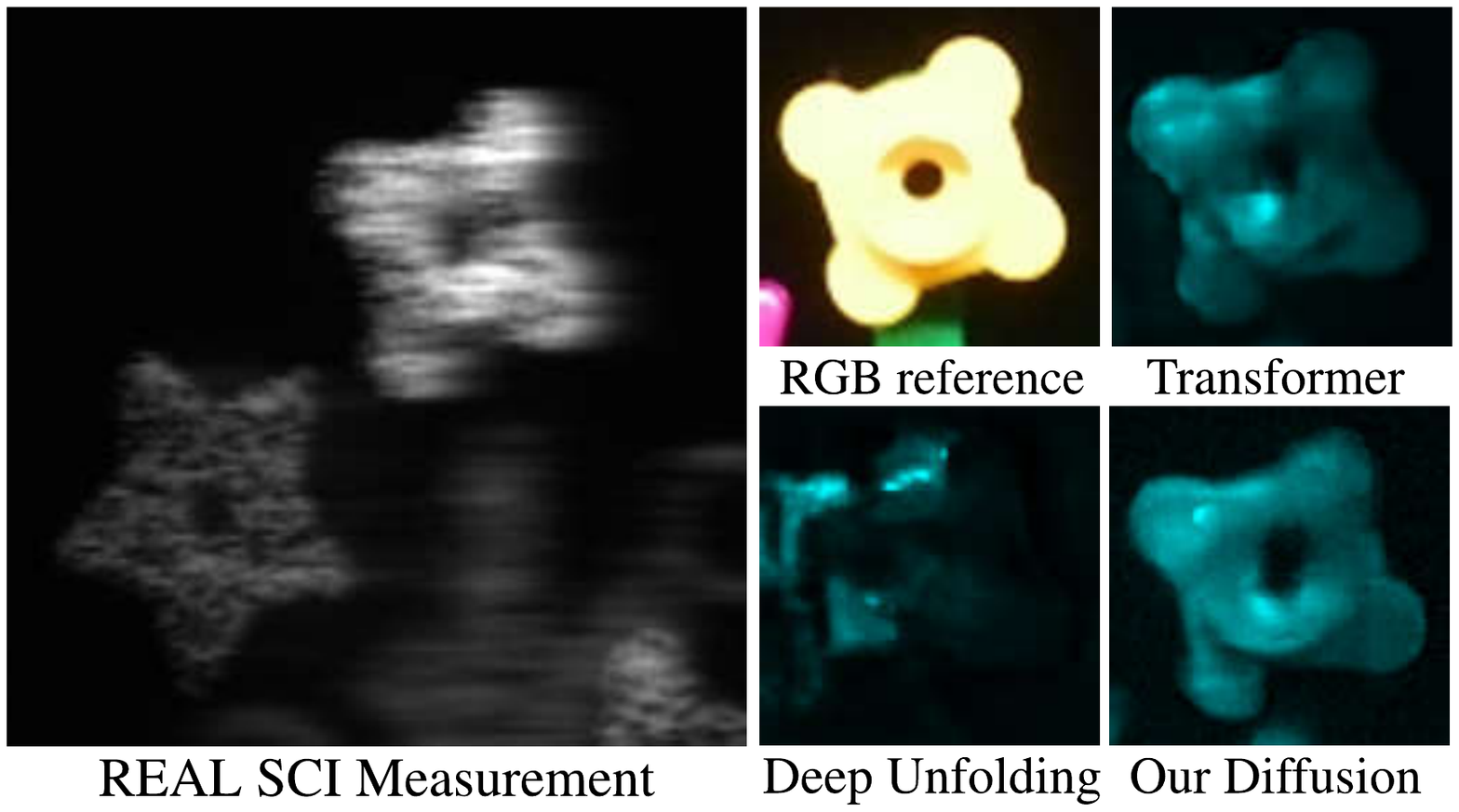}}
	\vspace{-9mm}
	\caption{Comparison of Transformer (MST~\cite{mst}), Deep Unfolding (TSA-Net~\cite{tsa_net}), and the proposed DiffSCI for real SCI reconstruction. The RGB image from the same scene serves as the reference. DiffSCI can reconstruct some unsampled and compressed scene contents by rethinking SCI through the generative diffusion model.} %
	  \vspace{-6mm}
	\label{fig_teaser}
\end{figure}

Given the ill-posed nature of SCI reconstruction as an inverse problem, existing methods still face several key challenges in accurately reconstructing certain aspects. For instance, inadequately illuminated regions or areas with sharp edges remain problematic as shown in Fig.~\ref{fig_teaser}. The underlying reason may be that insufficient sampling occurred in the above areas, then the reconstruction algorithm may not be able to accurately recover the detail information. Moreover, contemporary end-to-end (E2E) models~\cite{mi_3,tsa_net,hdnet,lambda}, while processing both two-dimensional measurements and three-dimensional MSIs maps, may inadvertently lose crucial high-dimensional information due to necessary dimensionality reduction. And current unsupervised methods also fail to achieve satisfactory results. Furthermore, the performance of the reconstruction on real-world datasets frequently deviates from the ideal, primarily attributable to discrepancies between the training dataset and the novel, unseen testing images, as evidenced in Fig.~\ref{fig_teaser}.

The diffusion model~\cite{nichol2021improved,conditionalddpm,kawar2022denoising,dhariwal2021diffusion} has demonstrated notable proficiency in generating content from RGB images~\cite{zhu2023denoising}. Leveraging its generative capacity to address challenging-to-reconstruct segments holds promise for enhancing MSIs SCI results~\cite{ddpm,ddim,conditionalddpm,anderson1982reverse,chung2022diffusion}.
Nonetheless, two significant challenges must be confronted:
\textbf{(i)} MSIs lack a substantial amount of training data for diffusion models compared to RGB images. Given the extensive band spectrum of MSIs, the temporal and GPU resources required for training would be significantly amplified. Consequently, training a diffusion model directly on MSIs proves to be a formidable task.
\textbf{(ii)} While utilizing a pre-trained diffusion model is a potential approach, current models are primarily trained on large RGB datasets, which inherently involve only three channels. In contrast, most MSIs encompass numerous bands, and the task of SCI reconstruction involves decoding a complete spatial-spectral MSI from a single measurement. This presents a notably distinct image restoration task with input and output dimensions that differ significantly. Consequently, the direct application of diffusion models to MSI reconstruction proves to be a non-trivial endeavor. 

Plug-and-Play (PnP)~\cite{pnp_1,pnp_2,self,yuan2021plug,pnpcassi,chan2016plug} framework incorporates pre-trained denoising networks into traditional model-based methods, due to its interpretability of the principles underlying SCI and its flexibility across different SCI systems, has emerged as one of the most predominant reconstruction techniques in the current scenario.
Therefore, we thought of using PnP framework to apply the pre-trained diffusion model based on massive RGB images as denoiser to the reconstruction of MSIs. However, there are four key challenges to embedding the diffusion model into MSIs at present. \textbf{(i)} Existing diffusion models are primarily applied to RGB images with three spectral bands, whereas MSIs typically involve dozens of spectral bands, MSIs cannot be fed directly into a diffusion model. \textbf{(ii)} There exists a spectral connection among the bands of MSIs, and many existing denoisers trained on RGB do not have a good grasp of this connection.
\textbf{(iii)} The wavelength range of RGB images is much smaller than that of MSIs, making wavelength mismatch issues inevitable. This discrepancy could significantly impact the performance of the diffusion model.
\textbf{(iv)} The sampling time required by the diffusion model in RGB images is already substantial. For our MSIs problem, the time required will be even greater.
In order to address these challenges, this paper makes the following contributions:

\begin{itemize}
\item Initially, the proposed DiffSCI leverages a diffusion model trained on a substantial corpus of RGB images for multispectral SCI reconstruction through the PnP framework, harnessing its generative potential to enhance SCI restoration outcomes. This is the first attempt to fill the research gap to fuse the diffusion model into the PnP framework for multispectral SCI.

\item
Acknowledging the inherent spectral band correlations in MSIs that are not present in RGB images, we embark on a comprehensive modeling of spectral correlation.

\item We introduce a method to address the inevitable issue of wavelength mismatch, given the broader spectral range of MSIs compared to RGB images.

\item We implement an accelerated strategy to get the analytic solution of the data subproblem within DiffSCI, which improves the convergence rate and reconstruction quality.

\end{itemize}

We validate DiffSCI through experiments on simulated and real datasets. Comparative assessments with state-of-the-art methods confirm DiffSCI's superior efficiency in restoring MSIs, as demonstrated by visual examples in Fig.~\ref{fig_teaser}.

\vspace{-2mm}
\section{Background}
\label{sec:related}
\vspace{-1mm}
\subsection{Degradation Model of CASSI}
\begin{figure*}[t]
	\begin{center}
		\begin{tabular}[t]{c} \hspace{-4mm}
                \includegraphics[width=1.01\textwidth]{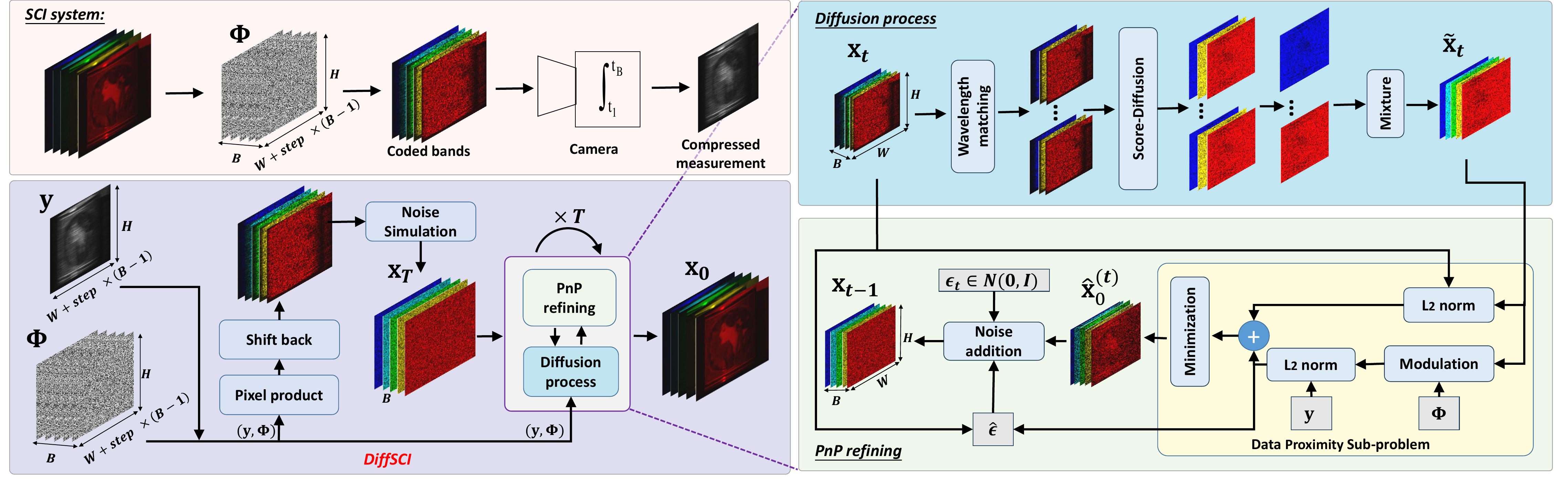}
		\end{tabular}
	\end{center}
	\vspace*{-9mm}
	\caption{\small \textbf{Top Left:} Obtaining 2D measurements $\mathbf{y}$ of 3D MSI through the SCI system with mask $\mathbf{\Phi}$. \textbf{Bottom Left:} DiffSCI generates desired reconstructed MSI ($\mathbf{x}_0$) with $\mathbf{y}$ and $\mathbf{\Phi}$ through reverse diffusion and PnP framework. \textbf{Right:} Integrating diffusion model with PnP method with wavelength matching (WM) method as a module of our DiffSCI method. }
	\label{fig:diffsci}
	\vspace{-6mm}
\end{figure*}

\vspace{-2mm}
In Coded Aperture Snapshot Spectral Compressive Imaging (CASSI) systems~\cite{sci_2,tsa_net,gehm2007single}, two-dimensional measurements $\mathcal{Y} \in \mathbb{R}^{H \times (W +d \times (B-1))}$ can be modulated from three-dimensional MSI $\mathcal{X} \in \mathbb{R}^{H \times W \times B}$ as shown in Fig.~\ref{fig:diffsci}, where $H, W, d$ and $B$ denote the MSI's height, width, shifting step and total number of wavelengths. 
As \cite{dauhst,admm-net}, we denote the vectorized measurement $\mathbf{y} \in \mathbb{R}^n$ with $n = H(W+d(B - 1))$. Then, given vectorized shifted MSI $\mathbf{x} \in \mathbb{R}^{nB}$ and mask $\mathbf{\Phi} \in \mathbb{R}^{n \times nB}$, the degradation model can be formulated as:
\begin{equation}
    \mathbf{y} = \mathbf{\Phi} \mathbf{x} + \mathbf{n},
\end{equation}
 where  $\mathbf{n}  \in \mathbb{R}^{n}$ represents the noise on measurement. SCI reconstruction is to obtain $\mathbf{x}$ from the captured $\mathbf{y}$ and the pre-set $\mathbf{\Phi}$ using a reconstruction algorithm~\cite{Tropp07ITT,Donoho06ITT,jalali2019snapshot}.

\subsection{Denoising Diffusion Probabilistic Models}
\vspace{-2mm}
Diffusion model includes two processes: forward process and reverse process. The forward process is to continuously add Gaussian noise to the clean image ($\mathbf{x}_0$) and eventually turn the initial image into pure Gaussian noise. Thus sampling $\mathbf{x}_t$ at any given timestep $t$ can be formulated as~\cite{ddpm}:
\begin{equation}
    \mathbf{x}_t = \sqrt{\Bar{\alpha}_t}\mathbf{x}_0 + \sqrt{1-\Bar{\alpha}_t}\epsilon,
\end{equation}
where $\alpha_t = 1 - \beta_t$, $\Bar{\alpha}_t = \prod_{k=1}^{t} \alpha_k$, $\epsilon \sim \mathcal{N}(0,\mathbf{I})$ and ${\beta_t}$ is a gradually increasing arithmetic sequence.
The reverse process is to gradually restore a clean image from Gaussian noise. One reverse step of Denoising Diffusion Probabilistic Models (DDPM) is~\cite{ddpm}:
\begin{equation}
    \mathbf{x}_{t-1} = \frac{1}{\sqrt{\alpha_t}}(\mathbf{x}_t - \frac{\beta_t}{\sqrt{1-\Bar{\alpha}_t}}\epsilon_{\theta}(\mathbf{x}_t,t))+\sqrt{\beta_t}\epsilon_t,
    \label{eq:DDPM_reverse}
\end{equation}
where $\epsilon_{\theta}(\mathbf{x}_t,t)$ is the noise predicted by the network at $t_{th}$ step and $\epsilon_t$ is standard Gaussian noise. Briefly, DDPM can be interpreted as a process of gradually subtracting the predicted noise from $\mathbf{x}_t$ to restore a clean image $\mathbf{x}_0$.

\subsection{Score-based Diffusion Model}
\vspace{-2mm}
Compared to DDPM, the score-based model can use methods like Langevin dynamics for more efficient sampling~\cite{song2020score}, and at the same time learn the data distribution (i.e., score function) under various noise levels, thus acquiring more training signals. This could help to improve the performance of the model.
The forward process can also be described in the form of a Stochastic Differential Equation (SDE):
\begin{equation}
    \mathbf{dx} = f(\mathbf{x},t)\mathbf{d}t + g(t)\mathbf{d}w,
\label{eq:SDE}
\end{equation}
where $\mathbf{d}w$ is infinitesimal white noise, $f(\cdot, t)$ is a vector function called the drift coefficient, and $g(\cdot, t)$ is a real-valued function called the diffusion coefficient.
The reverse process can be written as:
\begin{equation}
    \mathbf{d}\mathbf{x} = [f(\mathbf{x},t) - g^2(t)\nabla_\mathbf{x}\log p_t(\mathbf{x})]\mathbf{d}t+g(t)\mathbf{d}w,
    \label{equ:ddim}
\end{equation}
 where $p_t(\mathbf{x})$ is terminal distribution density~\cite{anderson1982reverse}, and the only unknown part $\nabla_\mathbf{x}logp_t(\mathbf{x})$ can be predicted through a score-based model $s_\theta(\mathbf{x},t)$~\cite{song2019generative,hyvarinen2005estimation}.

\subsection{Denoising Diffusion Implicit Models}
\vspace{-2mm}
In order to accelerate the reverse diffusion process, Denoising Diffusion Implicit Models (DDIM)  
generates new samples with a non-Markovian process. At each step, the model computes a denoised version of the image and then mixes this denoised version with some noise to generate the image for the next step. This process allows for more efficient estimation and sampling of multiple future states within the same time step, thus improving sampling efficiency and saving time.
Therefore, Eq.~\eqref{eq:DDPM_reverse} can be rewritten as:
 \begin{equation}
 \begin{aligned}
\mathbf{x}_{t-1}=&\sqrt{\bar{\alpha}_{t-1}}\left(\frac{\mathbf{x}_{t}-\sqrt{1-\bar{\alpha}_{t}} \mathbf{\epsilon}_\theta(\mathbf{x}_{t}, t)}{\sqrt{\bar{\alpha}_{t}}}\right)\\
+&\sqrt{1-\bar{\alpha}_{t-1}-{\sigma^2_{{\eta}_t}}} \cdot \mathbf{\epsilon}_\theta(\mathbf{x}_{t}, t)+\sigma_{{\eta}_t} \mathbf{\epsilon}_{t},
\label{equ:ddim}
\end{aligned}
 \end{equation}
the term inside the first bracket can be treated as denoised image $\Tilde{\mathbf{x}}_t$ predicted via current $\mathbf{x}_t$, $\sigma_{\eta_t}$ controls randomness.

\subsection{Conditional Diffusion Model}
\vspace{-2mm}
In the context of conditional generation tasks, we are presented with a condition $\mathbf{y}$, and our objective is to optimize the probability of $p(\mathbf{x}|\mathbf{y})$. Applying Bayes' theorem, we can rewrite Eq.~\eqref{equ:ddim} as~\cite{song2020score}:
\begin{equation}
\scalebox{0.9}{$
    \mathbf{d}\mathbf{x} = [f(\mathbf{x},t)-g^2(t)\nabla_\mathbf{x}(\log p_t(\mathbf{x})+\log p_t(\mathbf{y}|\mathbf{x}))]\mathbf{d}t+g(t)\mathbf{d}w, $}
    \label{equ:conditional}
\end{equation}
where the unconditionally pre-trained diffusion model achieves conditional generation by adding a classifier. So that, given Eq.~\eqref{equ:conditional}, one step of reverse sampling under conditional circumstances can be accomplished by first taking one reverse sampling step in the unconditional diffusion model, and then merging it with the conditional constraint.

\section{Proposed Method}
\label{sec:method}
\vspace{-1mm}
\subsection{Problem Definition and Solution}
\vspace{-2mm} 
Diffusion-based methods could theoretically recover the details of dark areas better through their powerful generative ability~\cite{low_light1,low_light2}. Unfortunately, the existing diffusion-based methods are mostly designed for RGB images in which the input and output are with three channels, while the task of SCI reconstruction involves decoding a complete multi-band MSI from a single-band measurement. Meanwhile, limited by the inadequate datasets of MSI and high dimension of the data, resource consumption required for retraining diffusion model on MSIs is high. To leverage the generative power of diffusion models and thus compensate for the shortcomings of current methods, our idea is to insert the pre-trained diffusion model on RGB images as a denoiser into the PnP framework to accomplish SCI reconstruction.

There are now four key problems: 
\textbf{(i)} How can diffusion models, trained on RGB images, be effectively applied to MSIs? \textbf{(ii)} How does one capture spectral correlation in MSIs that do not exist in RGB images? \textbf{(iii)} What strategies mitigate wavelength mismatching arising from inconsistencies between MSI and RGB wavelengths? \textbf{(iv)} How can fast and efficient sampling be achieved for MSIs with numerous bands?
To address the above issues, we proposed the DiffSCI method with three modules: Denoising Diffusion PnP-SCI Model, Diffusion Adaptation for MSI, and Acceleration Algorithm. See Fig.~\ref{fig:diffsci} for an overall view.

\subsection{Denoising Diffusion PnP-SCI Model}
\vspace{-2mm}
 The inversion problem of SCI can be modeled as:
\begin{equation}
\vspace{-2mm}
      \hat{\mathbf{x}} = \mathop{\arg\min}_\mathbf{x} \frac{1}{2} \|\mathbf{y} - \mathbf{\Phi} \mathbf{x}\|^2 + \lambda \mathcal{P}(\mathbf{x}), 
      \label{equ:HQS_1}
\end{equation}
where $\mathcal{P}(\mathbf{x})$ denotes diffusion MSI prior, $\lambda$ is a trade-off parameter. By adopting the half-quadratic splitting (HQS)~\cite{geman1995nonlinear} algorithm and introducing an auxiliary variable $\mathbf{z}$, Eq.~\eqref{equ:HQS_1} can be solved by iteratively solving following two subproblems:
\begin{align}
    & \mathbf{x}_{k+1}=\mathop{\arg\min}_{\mathbf{x}}\|\mathbf{y}-\mathbf{\Phi} \mathbf{x}\|^2  + \mu \| \mathbf{x} - \mathbf{z}_k \| ^2, \label{equ:HQS_3}\\
    & \mathbf{{z}}_{k+1}=\mathop{\arg\min}_{\mathbf{z}} \frac{\mu}{2} \|\mathbf{z} - \mathbf{x}_{k+1}\|^2 + \lambda \mathcal{P}(\mathbf{z}) \label{equ:HQS_4}.
\end{align}

\noindent\textbf{Closed-form Solution to Data Subproblem.} In CASSI system, $\mathbf{\Phi}^T \mathbf{\Phi}$ is a diagonal matrix~\cite{pnpcassi,dauhst}, so that by
using matrix inversion theorem (Woodbury matrix identities), the closed-form solution of Eq.~\eqref{equ:HQS_3} can be easily found with fast operation guarantee~\cite{daubechies2004iterative}:
\begin{equation}
\mathbf{x}_{k+1} = \mathbf{z}_k + \mathbf{\Phi}^T[\mathbf{y} - \mathbf{\Phi}\mathbf{z}_k]\oslash[Diag(\mathbf{\Phi} \mathbf{\Phi}^T) + \mu], 
\label{equ:HQS_6}
\end{equation}
where $Diag(\cdot)$ extracts the diagonal elements of the ensured matrix, $\oslash$ is the element-wise division of Hadamard division. 

\noindent\textbf{Diffusion Models as Generative Denoiser Prior.}
Unlike conventional denoisers, diffusion models possess powerful generative capabilities~\cite{deja2022analyzing}. To utilize this generative capability, our DiffSCI model explores diffusion as the generative denoiser prior as shown in Fig.~\ref{fig:diffsci} to address hard-to-recover parts of SCI reconstruction, such as low-light and sharp edges. We firstly establish the correlation between Eq.~\eqref{equ:HQS_4} and diffusion model.
Let $\mathbf{x}_k^{(b)}$ be a three-channel image corresponding to $b_{th}$ band of MSI $\mathbf{x}_k$, from Eq.~\eqref{equ:HQS_4} we have:
\begin{equation}
    {\mathbf{z}^{(b)}_{k+1}} = \mathop{\arg\min}_{\mathbf{z}^{(b)}} \frac{1}{2(\sqrt{\lambda/\mu})^2}\| \mathbf{z}^{(b)} - \mathbf{x}_{k+1}^{(b)} \|+ \mathcal{P}(\mathbf{z}^{(b)}),
    \label{equ:HQS_7}
\end{equation}
where $\mathbf{z}^{(b)}_{k+1}$ can be treated as clean image from noisy image $\mathbf{x}_{k+1}^{(b)}$ with noise level $\Bar{\sigma}_t = \sqrt{\frac{1-\Bar{\alpha}_t}{\Bar{\alpha}_t}}$. Letting $\Bar{\sigma}_t = \sqrt{\lambda/\mu}$, with $\nabla_\mathbf{x} \mathcal{P}(\mathbf{x}) = -\nabla_\mathbf{x} \log p(\mathbf{x}) = -\mathbf{s}_\theta(\mathbf{x})$ ~\cite{zhu2023denoising}, Eq.~\eqref{equ:HQS_7} can be rewritten as:
\begin{align}
\vspace{-2mm}
    \mathbf{z}^{(b)}_{k+1} &\approx \mathbf{x}_{k+1}^{(b)} + \frac{1-\Bar{\alpha}_t}{\Bar{\alpha}_t}\mathbf{s}_\theta(\mathbf{x}_{k+1}^{(b)},t).
\end{align}
Hence, we can perceive $\mathbf{z}^{(b)}_{k+1}$ as the clean three-channel image $\Tilde{\mathbf{x}}^{(b)}_{k+1}$ reversed from $\mathbf{x}_{k+1}^{(b)}$. 

\subsection{Diffusion Adaptation for MSI}
\vspace{-2mm}
Applying an RGB pre-trained denoising diffusion model directly to MSI would cause issues such as band number mismatching, insufficient spectral correlation, and wavelength mismatching. This section will investigate these problems.

\begin{figure}
    \centering
    \begin{subfigure}[t]{0.45\linewidth}
    \begin{overpic}[height=\linewidth, width=\linewidth]{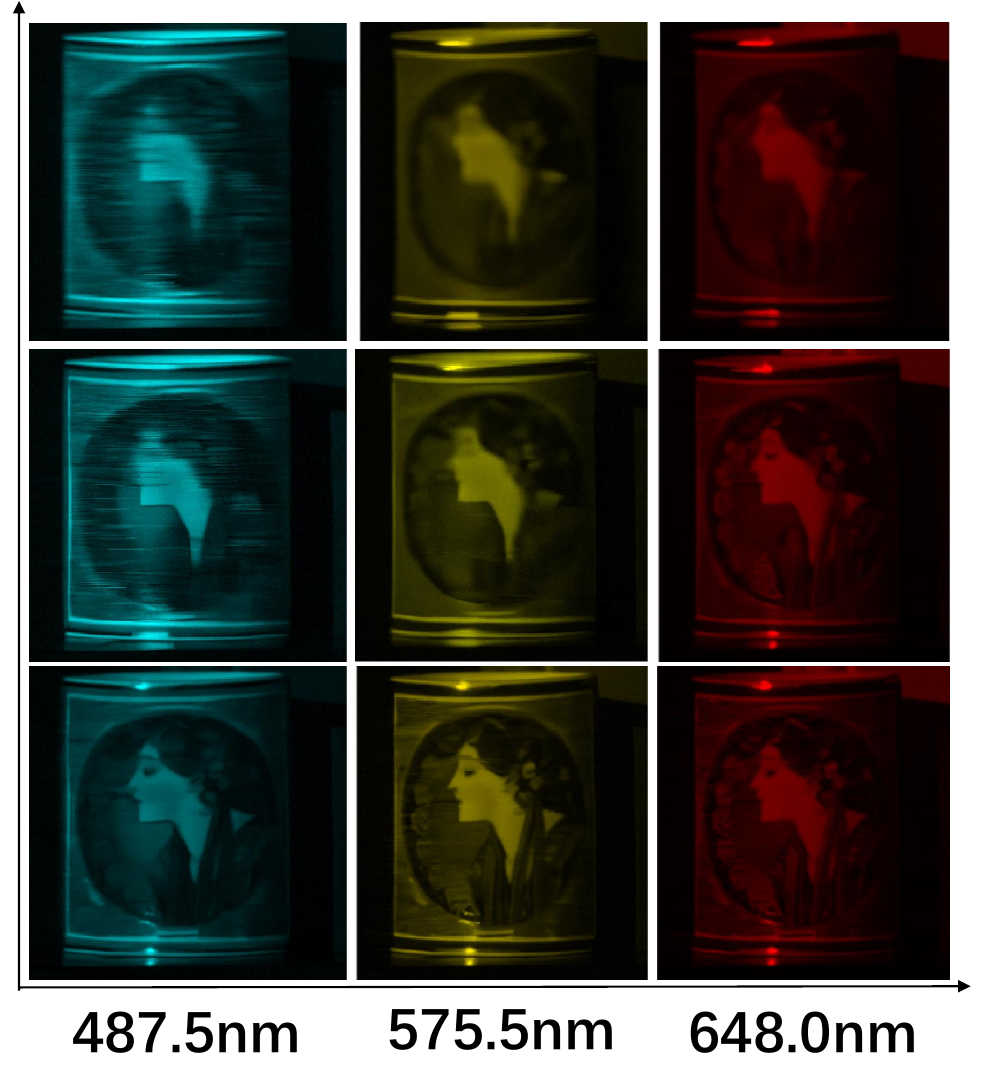}
\put(-3,82){\color{black}{\scriptsize a}}
\put(-3,50){\color{black}{\scriptsize b}}
\put(-3,20){\color{black}{\scriptsize c}}
\end{overpic}
\end{subfigure}
\begin{subfigure}[t]{0.45\linewidth}
\begin{overpic}[height = \linewidth, width=1.2\linewidth]{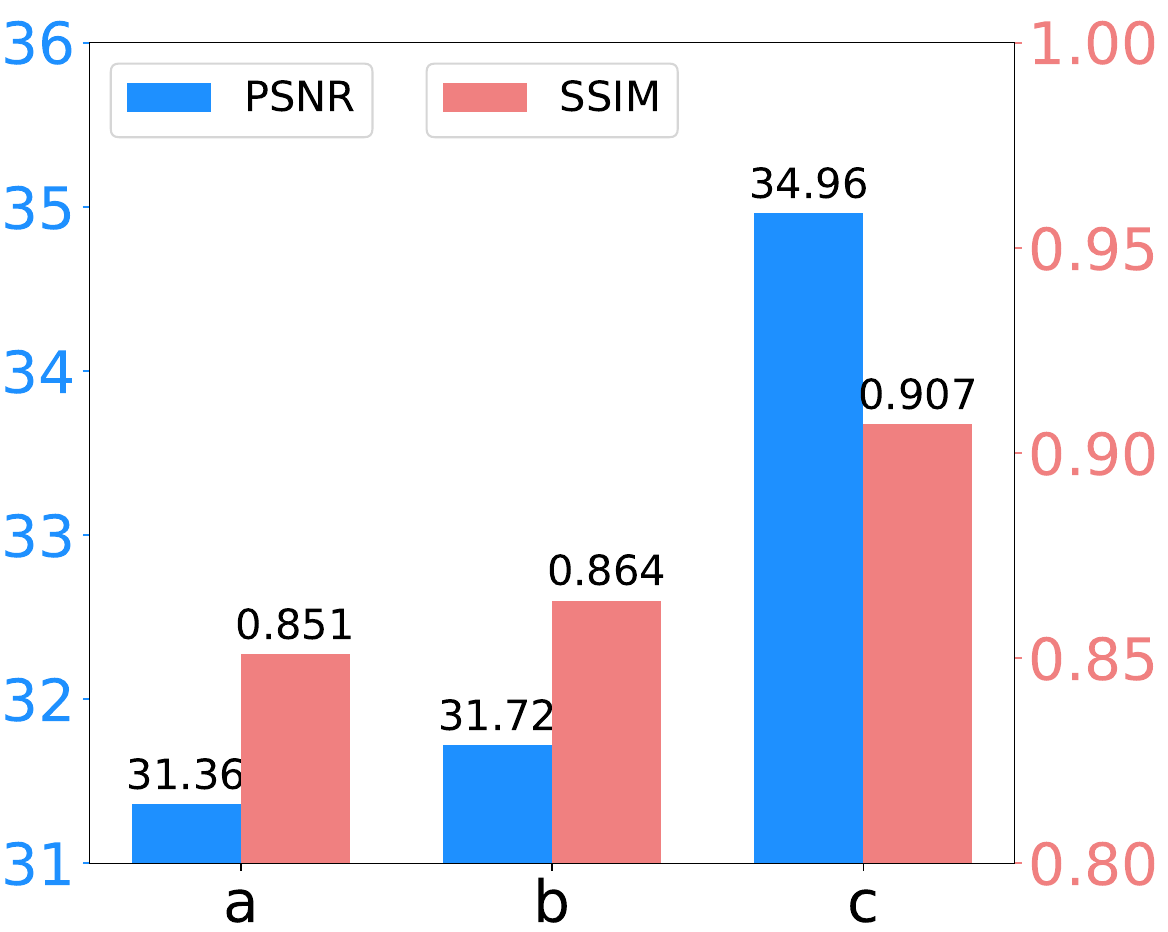}
\end{overpic}
\end{subfigure}
\vspace{-3mm}
    \caption{Visual effects and PSNR/SSIM presentation of (a) independently selecting non-overlapping bands method, (b) spectral correlation modeling, and (c) wavelength matching method of $Scene$ 1 of 3 (out of 28) spectral channels.}
    \label{fig:wm_1}
    \vspace{-5mm}
\end{figure}

\noindent\textbf{Spectral Correlation Modeling.}
MSIs exhibit spectral correlation between neighboring bands, denoted as $[B_{i-1}, B_i, B_{i+1}]$. However, conventional PnP methods treat each band independently, performing denoising operations as $R_i = D(B_i)$, thereby neglecting this inherent spectral correlation. One approach to address this correlation is to partition the MSIs into distinct, non-overlapping bands,
\begin{equation}
\scalebox{0.95}{$
    C_k = [B_{i-1}, B_i, B_{i+1}], C_{k+1} = [B_{i+2}, B_{i+3}, B_{i+4}], $}
\end{equation}
but it just models the part spectral correlation which may cause pixel jump between $B_{i+1}$ and $B_{i+2}$.
Here, to model the spectral correlation, for each band reconstruction, we extract adjacent bands for combination,
\begin{equation}
C_k = [B_{i-1}, B_i, B_{i+1}], C_{k+1} = [B_{i}, B_{i+1}, B_{i+2}],
\end{equation}
the combined representation serves as the input for the diffusion model. Subsequently, the corresponding band from the output is selected as the recovered band $R_i$ for the MSIs,
\begin{equation}
R_i = D(C).
\end{equation}
\emph{Quality Comparison:}
The quality ($Q$) of the reconstructed MSIs obtained through the spectral correlation modeling method is significantly superior compared to individually selecting non-overlapping bands as shown in Fig.~\ref{fig:wm_1}, i.e.,
\begin{equation}
Q(D(C)) > Q(D([B_{i-1}, B_i, B_{i+1}])).
\end{equation}

\noindent\textbf{Wavelength Matching.}
Based on previous experiments illustrated in Fig.~\ref{fig:wm_1}, it was observed that the reconstruction performance of forward bands was significantly inferior compared to later bands. Analyzing the \emph{Spectral Bands and Range} within the simulated dataset revealed the division of MSIs into 28 spectral bands spanning from 450nm to 720nm,
\begin{equation}
   \text{Bands} = \{B_i\}_{i=1}^{28}, \quad \lambda(B_i) \in [450, 720].
\end{equation}
While the spectral bands of the RGB image are only a subset of these, i.e.,
\begin{equation}
 \scalebox{0.9}{$
\lambda(\text{RGB}) = \{660, 520, 450\}\subset [453, 720], \text{RGB} \subset \text{MSIs}. $}
\end{equation}
Hence, establishing wavelength matching (WM) between MSIs and RGB images is imperative. In the context of recovering bands with wavelengths significantly distant from RGB images, our DiffSCI method integrates them with two bands featuring matched wavelengths, thereby mitigating interference arising from wavelength mismatching,
\begin{equation}
\text{WM}(B_i) = \text{Merge}(B_i, B_{i+n}, B_{i+m}).
\end{equation}
\emph{Enhanced Metrics:} Experimental findings demonstrate significant improvement in both PSNR and SSIM when employing this approach in conjunction with previous spectral correlation modeling methods, as illustrated in Fig.~\ref{fig:wm_1}.

\begin{figure}[!t]
\centering
\begin{subfigure}{0.45\linewidth}
\begin{overpic}[height=\linewidth, width=1.05\linewidth]{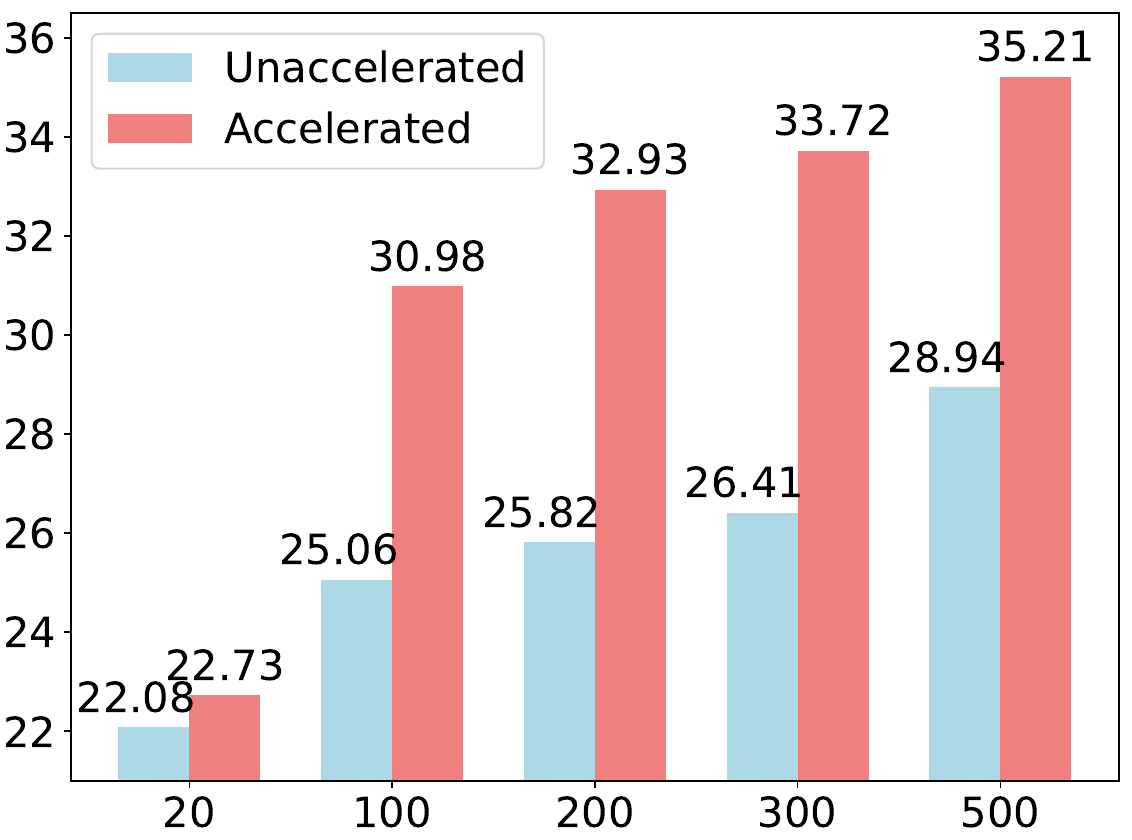}
\put(28,-5){\color{black}{\scriptsize Sampling steps}}
\put(-7,40){\rotatebox{90}{\color{black}{\scriptsize PSNR/db}}}
\end{overpic}
\end{subfigure}
\hspace{4mm}
\begin{subfigure}{0.45\linewidth}
\begin{overpic}[height=\linewidth, width=1.05\linewidth]{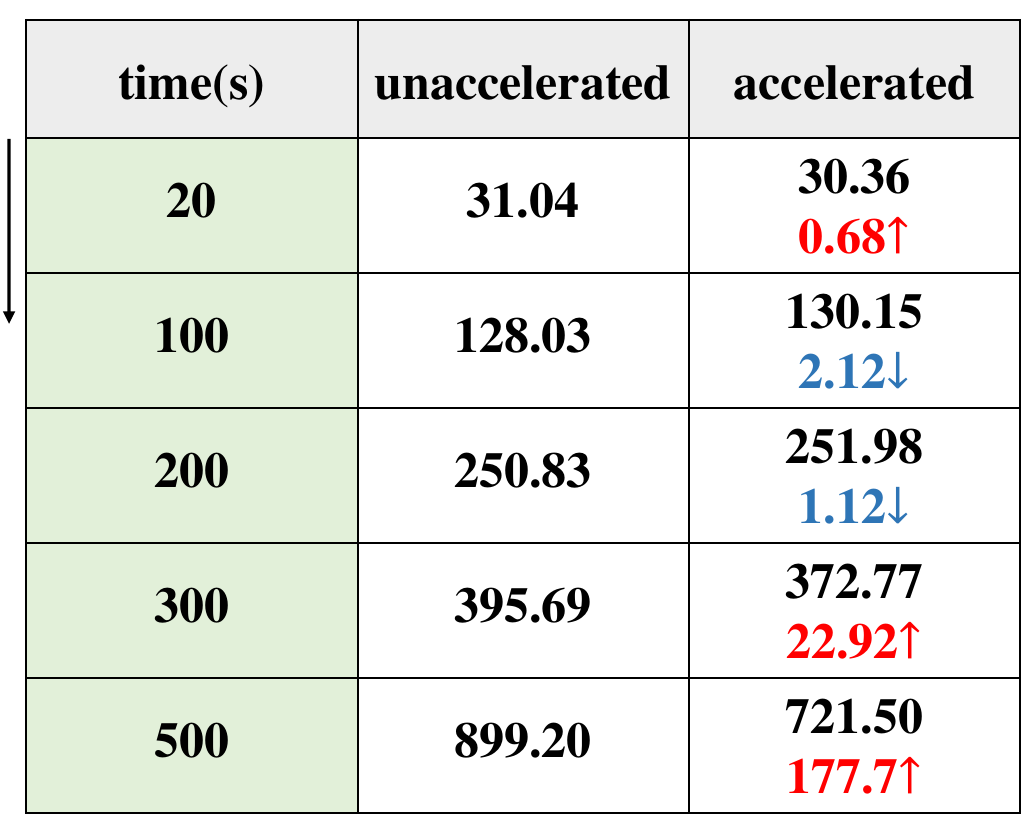}
\put(-4,62){\rotatebox{90}{\color{black}{\scriptsize steps}}}
\put(28,-5){\color{black}{\scriptsize Time of each step}}
\end{overpic}
\end{subfigure}
\vspace{-1mm}
    \caption{Effect of sampling steps and acceleration algorithm on scene5 of simulation dataset on PSNR and time.}
    \label{fig:acc}
    \vspace{-6mm}
\end{figure}

\vspace{-2mm}
\subsection{Acceleration Algorithm}
\vspace{-2mm}
Motivated by the fact that the sampling process of diffusion model is time-consuming and unconditional, we employ an acceleration algorithm to achieve faster and more efficient sampling.
As mentioned in Eq.~\eqref{equ:HQS_6}, current methods usually calculate residuals by ($\mathbf{y} - \Phi \mathbf{z}_k$), which only uses information about the current $\mathbf{z}_k$ for iterative updates. As a result, this approach leads to slow convergence speed and fails to effectively address the issue of data proximity.

On this basis, we introduce a variable $\mathbf{y}_1$, which can be defined as $ \mathbf{y}_1 = \mathbf{y}_1 + (\mathbf{y} - \mathbf{\Phi} \mathbf{z}_k)$, can be treated as the accumulation of residuals and calculate residuals by calculating ($\mathbf{y}_1 - \mathbf{\Phi} \mathbf{z}_k$) iteratively. On the one hand, $\mathbf{y}_1$ can be used to incorporate more residual information for updating $\mathbf{z}$, thereby improving reconstruction quality. On the other hand, a form similar to Nesterov acceleration~\cite{nesterov} is employed to expedite the convergence speed.

\noindent\textbf{Accumulation of Residuals.}
Since $\mathbf{y}_1$ is updated at each iteration, it contains all the residual information from previous iterations. This means that when we update $\mathbf{z}$ using $\mathbf{y}_1$, we are effectively utilizing information from all previous iterations, not just the most recent one.

\noindent\textbf{Methods Pertaining to Nesterov-Type Acceleration.} The closed-form solution Eq.~\eqref{equ:HQS_6} can be rewritten as:
 \begin{equation}
        \mathbf{x}_{k+1} = \mathbf{z}_k + \mathbf{\Phi}^T[\mathbf{y}_1 - \mathbf{\Phi}\mathbf{z}_k]\oslash[Diag(\mathbf{\Phi} \mathbf{\Phi}^T) + \mu]   
 \end{equation}
 \vspace{-4mm}
\begin{equation}
=\mathbf{z}_k+ \mathbf{\Phi}^T [\sum_{i=1}^{k} (\mathbf{y} - \mathbf{\Phi} \mathbf{z}_{i}) - \mathbf{\Phi}\mathbf{z}_k] \oslash[Diag(\mathbf{\Phi} \mathbf{\Phi}^T) + \mu].  \nonumber 
 \end{equation}
Thus, we can approximate that $\mathbf{x}_{k+1}$ is derived from $\sum_{i=1}^{k} (\mathbf{y} - \mathbf{\Phi} \mathbf{z}_{i})$ and $\mathbf{z}_{k}$, resembling Nesterov's acceleration concept. This enhances the efficacy of the data fidelity term and accelerates the overall convergence rate of the algorithm, as evidenced by experimental comparisons in Fig.~\ref{fig:acc}.

Meanwhile, we define $guidance\_scale$ ($sc$) as the iterative step size as the data subproblem and test the effect of different $sc$ on the results, which are shown in Fig.~\ref{fig:guidance}.

\begin{algorithm}[!t]
    \renewcommand{\algorithmicrequire}{\textbf{Require:}}
	\renewcommand{\algorithmicensure}{\textbf{Output:}}
   \caption{DiffSCI sampling}
   \label{alg:diffsci}
    \begin{algorithmic}[1]
     \REQUIRE$\mathbf{s}_\theta$, $T$,$B$, $\mathbf{y}$, $\mathbf{\Phi}$, $\sigma_{n}$,  ${\{\bar\sigma_{t}\}_{t=1}^T}$, $\zeta$, $\lambda$
    \STATE{Initialize $\mathbf{x}_{T}\sim \mathcal{N}(\mathbf{0}, \mathbf{I})$, $\mathbf{y}_1 = 0$, pre-calculate $\rho_t \triangleq  {\lambda\sigma_{n}^2}/{\bar\sigma_{t}^2}$.}
    \FOR{$t=T$ {\bfseries to} $1$}
        \FOR{$b = 1$ {\bfseries to} $B$}
        \STATE{$\mathbf{x}_{t}^ {\scalebox{0.6}{$(b)$}}$ 
             = WM($\mathbf{B}_b$)\ \textcolor[rgb]{0.40,0.40,0.40}{\textit{ // wavelength mathcing method}}} 
        \vspace{1mm}
        \STATE{{$\Tilde{\mathbf{x}}_t^{\scalebox{0.6}{$(b)$}}=
        \frac{1}{\sqrt{\bar\alpha_t}}(\mathbf{x}_t^{\scalebox{0.6}{$(b)$}} + (1 - \bar\alpha_t)\mathbf{s}_\theta(\mathbf{x}_t^{\scalebox{0.6}{$(b)$}},t))$} \textcolor[rgb]{0.40,0.40,0.40}{\textit{//predict clean image from $\mathbf{x}_t^{\scalebox{0.6}{$(b)$}}$ with score based model}}}
        \ENDFOR 
    \STATE{Get $\Tilde{\mathbf{x}}_t$ \textcolor[rgb]{0.40,0.40,0.40}{\textit{// combination }}
    \STATE{$\mathbf{y}_1 = \mathbf{y}_1 + (\mathbf{y} - \mathbf{\Phi} \Tilde{\mathbf{x}}_t)$}\textcolor[rgb]{0.40,0.40,0.40}{\textit{ // calculate and accumulate residuals }}
    \STATE{$\hat{\mathbf{x}}_0^{(t)} = \Tilde{\mathbf{x}}_t + sc \cdot (\mathbf{y}_1 - \mathbf{\Phi} \Tilde{\mathbf{x}}_t) \oslash[Diag(\mathbf{\Phi} \mathbf{\Phi}^T) + \rho_t]$}\textcolor[rgb]{0.40,0.40,0.40}{\textit{ // acceleration for data subproblem}}
   
    
    \STATE{$ \hat{\mathbf{\epsilon}} = \frac{1}{\sqrt{1 - \bar{\alpha}_t}} (\mathbf{x}_t -\sqrt{\bar{\alpha}_t}\mathbf{\hat{x}}_{0}^{\scalebox{0.6}{$(t)$}} )$}
    \vspace{1mm}
    \STATE{$ \mathbf{\epsilon}_{t} \sim \mathcal{N}(\mathbf{0}, \mathbf{I})$}
    \STATE{$\mathbf{x}_{t-1}=\sqrt{\bar{\alpha}_{t-1}}\mathbf{\hat{x}}_{0}^{\scalebox{0.6}{$(t)$}}+ \sqrt{1 - \bar{\alpha}_{t-1}}(\sqrt{1-\zeta}\hat{\mathbf{\epsilon}} + \sqrt{\zeta}\mathbf{\epsilon}_{t})$  \textcolor[rgb]{0.40,0.40,0.40}{\textit{ // diffusion to $\mathbf{x}_{t-1}$ to finish one step sampling}}}}
    \ENDFOR 
\STATE {\bfseries return} $\mathbf{x}_0$
\end{algorithmic}
\end{algorithm}

\vspace{-1mm}
\subsection{DiffSCI Method}
\vspace{-1mm}
In DiffSCI, we embed diffusion model into SCI via PnP framework. To elaborate, we can rewrite it as:
\begin{align}
    &\mathbf{x}^{(b)}_t \xleftarrow{\text{WM}(B_b)} \mathbf{x}_t\label{equ:algorithm1},
    \\
    &\Tilde{\mathbf{x}}_t^{(b)} = \mathop{\arg\min}_{\mathbf{z}^{(b)}} \frac{1}{2 \Bar{\sigma}_t^2}\| \mathbf{z}^{(b)} - \mathbf{x}^{(b)}_t \|+ \mathcal{P}(\mathbf{z}^{(b)})\label{equ:algorithm2},
    \\
    &\Tilde{\mathbf{x}}_t \xleftarrow{\text{combination }} \Tilde{\mathbf{x}}_t^{(b)}\label{que:algorithm3},
    \\
    &{\hat{\mathbf{x}}_0}^{(t)} =\mathop{\arg\min}_{\mathbf{x}} \| \mathbf{y} - \mathbf{\Phi}(\mathbf{x})\|^2 + \rho_t \| \mathbf{x} - \Tilde{\mathbf{x}}_t\|^2\label{equ:algorithm4},
    \\
    & \mathbf{x}_{t-1} \leftarrow{{\hat{\mathbf{x}}_0}^{(t)}},
    \label{equ:algorithm5}
    \vspace{-2mm}
\end{align}
where $\rho_t = \lambda(\sigma_n/\bar{\sigma}_t)^2$, $\mathbf{x}_t$ is noisy MSI at timestep $t$, $\mathbf{x}_t^{(b)}$ denotes the three-channel image at timestep $t$ obtained by WM($B_b$) from $\mathbf{x}_t$, $\Tilde{\mathbf{x}}_t^{(b)}$ is noiseless three-channel image of $\mathbf{x}_t^{(b)}$ and $\Tilde{\mathbf{x}}_t$ denotes noiseless MSI through combination.

\noindent\textbf{DiffSCI Sampling.}
According to previous discussion, the clean estimated MSI $\hat{\mathbf{x}}_0^{(t)}$ can be obtained from $\mathbf{x}_t$ with the condition $\mathbf{y}$. However, this estimation is not accurate, we can add noise and diffusion to timestep $t-1$ as Eq.~\eqref{equ:algorithm5}. $\hat{\mathbf{x}}_0^{(t)}$ with condition $\mathbf{y}$ can be firstly gotten, whose conditional distribution is $p(x|y)$, and estimated clean image can be used to calculate the noise with condition $\mathbf{y}$, which is $\hat{\mathbf{\epsilon}} = \frac{1}{\sqrt{1 - \bar{\alpha}_t}} (\mathbf{x}_t -\sqrt{\bar{\alpha}_t}\mathbf{\hat{x}}_{0}{(t)})$. Then, the diffusion expression like Eq.~\eqref{equ:ddim} is:
\begin{equation}
    \mathbf{x}_{t-1} = \sqrt{\bar{\alpha}_{t-1}}\hat{\mathbf{x}}_0^{(t)} + \sqrt{1-\bar{\alpha}_{t-1}-{\sigma^{2}_{{\eta}_t}}}\hat{\mathbf{\epsilon}}+\sigma_{{\eta}_t} \mathbf{\epsilon}_{t}.
    \label{equ:algorithm6}
\end{equation}
Based on previous experience~\cite{zhu2023denoising}, the noise term $\sigma_{{\eta}_t}$ could be set to 0, and hyperparameter $\zeta$ can be used to introduce noise to balance $\epsilon_t$ and $\hat{\epsilon}$, and Eq.~\eqref{equ:algorithm6} can be rewritten as:
\begin{equation}
    \mathbf{x}_{t-1} = \sqrt{\bar{\alpha}_{t-1}}\hat{\mathbf{x}}_0^{(t)}+\sqrt{1-\bar{\alpha}_{t-1}}(\sqrt{1-\zeta} \hat{\epsilon} + \sqrt{\zeta}\epsilon_t),
\end{equation}
where $\zeta$ controls the variance of the noise added at each step, when $\zeta = 0$, our method becomes a deterministic process.

Finally, we summarize the algorithm for DiffSCI-based MSI reconstruction in Algorithm \ref{alg:diffsci}. Further details regarding the model are presented in the supplementary materials.

\begin{table*}

	\caption{Comparisons between DiffSCI and SOTA methods on 10 simulation scenes (S1$\sim$S10). Category, PSNR (upper entry in each cell), and SSIM (lower entry in each cell) are reported. The best and second best results are highlighted in bold and underlined, respectively.}
 \vspace{-3mm}
	\newcommand{\tabincell}[2]{\begin{tabular}{@{}#1@{}}#2\end{tabular}}
	\centering
	\resizebox{0.99\textwidth}{!}
	{
		\centering
		\begin{tabular}{ccccccccccccccc}
			\bottomrule[0.15em]
			\rowcolor{lightgray}
			~~~~~Algorithms~~~~~
			&~~Category~~
                &~~Reference~~
			& ~~~~~S1~~~~~
			& ~~~~~S2~~~~~
			& ~~~~~S3~~~~~
			& ~~~~~S4~~~~~
			& ~~~~~S5~~~~~
			& ~~~~~S6~~~~~
			& ~~~~~S7~~~~~
			& ~~~~~S8~~~~~
			& ~~~~~S9~~~~~
			& ~~~~~S10~~~~~
			& ~~~~Avg~~~~
			\\
			\midrule
                \rowcolor{lightgray5}
			TwIST \cite{twist}
			& \tabincell{c}{Model}
                & TIP 2007
			&\tabincell{c}{25.16\\0.700}
			&\tabincell{c}{23.02\\0.604}
			&\tabincell{c}{21.40\\0.711}
			&\tabincell{c}{30.19\\0.851}
			&\tabincell{c}{21.41\\0.635}
			&\tabincell{c}{20.95\\0.644}
			&\tabincell{c}{22.20\\0.643}
			&\tabincell{c}{21.82\\0.650}
			&\tabincell{c}{22.42\\0.690}
			&\tabincell{c}{22.67\\0.569}
			&\tabincell{c}{23.12\\0.669}
			\\
               \midrule
               \rowcolor{lightgray5}
			GAP-TV \cite{gap_tv}
			& \tabincell{c}{Model}
                & ICIP 2016
			&\tabincell{c}{26.04\\0.817}
			&\tabincell{c}{21.66\\0.724}
			&\tabincell{c}{24.86\\0.732}
			&\tabincell{c}{30.51\\0.875}
			&\tabincell{c}{24.33\\0.778}
			&\tabincell{c}{25.11\\0.790}
			&\tabincell{c}{18.28\\0.730}
			&\tabincell{c}{23.94\\0.780}
			&\tabincell{c}{21.77\\0.732}
			&\tabincell{c}{23.08\\0.721}
			&\tabincell{c}{23.96\\0.768}
			\\
			  \midrule
                \rowcolor{lightgray5}
			DeSCI \cite{desci}
			& \tabincell{c}{Model}
               & TPAMI 2019
			&\tabincell{c}{28.38\\0.803}
			&\tabincell{c}{26.00\\0.701}
			&\tabincell{c}{23.11\\0.730}
			&\tabincell{c}{28.26\\0.855}
			&\tabincell{c}{25.41\\0.778}
			&\tabincell{c}{24.66\\0.764}
			&\tabincell{c}{24.96\\0.725}
			&\tabincell{c}{24.15\\0.747}
			&\tabincell{c}{23.56\\0.701}
			&\tabincell{c}{24.17\\0.677}
			&\tabincell{c}{25.27\\0.748}
			\\
                \midrule
                \rowcolor{rouse}
			$\lambda$-Net \cite{lambda}
			& \tabincell{c}{CNN\\(Supervised)}
               & ICCV 2019
			&\tabincell{c}{30.10\\0.849}
			&\tabincell{c}{28.49\\0.805}
			&\tabincell{c}{27.73\\0.870}
			&\tabincell{c}{37.01\\0.934}
			&\tabincell{c}{26.19\\0.817}
			&\tabincell{c}{28.64\\0.853}
			&\tabincell{c}{26.47\\0.806}
			&\tabincell{c}{26.09\\0.831}
			&\tabincell{c}{27.50\\0.826}
			&\tabincell{c}{27.13\\0.816}
			&\tabincell{c}{28.53\\0.841}
			\\
                \midrule
                \rowcolor{rouse}
			TSA-Net \cite{tsa_net}
			& \tabincell{c}{CNN\\(Supervised)}
               & ECCV 2020
			&\tabincell{c}{32.31\\0.894}
			&\tabincell{c}{31.03\\0.863}
			&\tabincell{c}{32.15\\0.916}
			&\tabincell{c}{37.95\\0.958}
			&\tabincell{c}{29.47\\0.884}
			&\tabincell{c}{31.06\\0.902}
			&\tabincell{c}{30.02\\0.880}
			&\tabincell{c}{29.22\\0.886}
			&\tabincell{c}{31.14\\0.909}
			&\tabincell{c}{29.18\\0.861}
			&\tabincell{c}{31.35\\0.895}
			\\
                \midrule
                \rowcolor{rouse}
			HDNet \cite{hdnet}
			& \tabincell{c}{Transformer\\(Supervised)}
               & CVPR 2022
			&\tabincell{c}{34.96\\0.937}
			&\tabincell{c}{35.64\\0.943}
			&\tabincell{c}{35.55\\0.946}
			&\tabincell{c}{41.64\\0.976}
			&\tabincell{c}{32.56\\0.948}
			&\tabincell{c}{34.33\\0.954}
			&\tabincell{c}{33.27\\0.928}
			&\tabincell{c}{32.26\\0.945}
			&\tabincell{c}{34.17\\0.944}
			&\tabincell{c}{32.22\\0.940}
			&\tabincell{c}{34.66\\0.946}
			\\
                \midrule
                \rowcolor{rouse}
			MST-L \cite{mst}
			& \tabincell{c}{Transformer\\(Supervised)}
               & CVPR 2022
			&\tabincell{c}{35.30\\0.944}
			&\tabincell{c}{36.13\\0.948}
			&\tabincell{c}{35.66\\0.954}
			&\tabincell{c}{40.05\\0.976}
			&\tabincell{c}{32.84\\0.949}
			&\tabincell{c}{34.56\\0.955}
			&\tabincell{c}{33.80\\0.930}
			&\tabincell{c}{32.74\\0.950}
			&\tabincell{c}{34.37\\0.944}
			&\tabincell{c}{32.63\\0.943}
			&\tabincell{c}{34.81\\0.949}
			\\
                \midrule
                \rowcolor{rouse}
			MST++ \cite{mst_pp}
			& \tabincell{c}{Transformer\\(Supervised)}
               & CVPR 2022
			&\tabincell{c}{\underline{35.57}\\ \underline{0.945}}
			&\tabincell{c}{\underline{36.22}\\ \underline{0.949}}
			&\tabincell{c}{37.00\\ \underline{0.959}}
			&\tabincell{c}{\textbf{42.86}\\ \underline{0.980}}
			&\tabincell{c}{\underline{33.27}\\ \underline{0.954}}
			&\tabincell{c}{\underline{35.27}\\ \underline{0.960}}
			&\tabincell{c}{34.05\\0.936}
			&\tabincell{c}{\underline{33.50}\\ \underline{0.956}}
			&\tabincell{c}{\underline{36.17}\\ \underline{0.956}}
			&\tabincell{c}{\textbf{33.26}\\ \underline{0.949}}
			&\tabincell{c}{\textbf{35.72}\\ \underline{0.955}}
			\\
                \midrule
                \rowcolor{rouse}
			CST-L+ \cite{cst}
			& \tabincell{c}{Transformer\\(Supervised)}
               & ECCV 2022
			&\tabincell{c}{\textbf{35.64}\\ \textbf{0.951}}
			&\tabincell{c}{\textbf{36.79}\\ \textbf{0.957}}
			&\tabincell{c}{\underline{37.71}\\ \textbf{0.965}}
			&\tabincell{c}{41.38\\ \textbf{0.981}}
			&\tabincell{c}{32.95\\ \textbf{0.957}}
			&\tabincell{c}{\textbf{35.58}\\\textbf{0.966}}
			&\tabincell{c}{\underline{34.54}\\ \textbf{0.947}}
			&\tabincell{c}{\textbf{34.07}\\ \textbf{0.964}}
			&\tabincell{c}{35.62\\ \textbf{0.959}}
			&\tabincell{c}{\underline{32.82}\\ \textbf{0.949}}
			&\tabincell{c}{\underline{35.71}\\ \textbf{0.960}}
                \\
                \midrule
                \rowcolor{light-yellow}
			DGSMP \cite{gsm}
			& \tabincell{c}{Deep Unfolding\\(Supervised)}
               & CVPR 2021
			&\tabincell{c}{33.26\\0.915}
			&\tabincell{c}{32.09\\0.898}
			&\tabincell{c}{33.06\\0.925}
			&\tabincell{c}{40.54\\0.964}
			&\tabincell{c}{28.86\\0.882}
			&\tabincell{c}{33.08\\0.937}
			&\tabincell{c}{30.74\\0.886}
			&\tabincell{c}{31.55\\0.923}
			&\tabincell{c}{31.66\\0.911}
			&\tabincell{c}{31.44\\0.925}
			&\tabincell{c}{32.63\\0.917}
			\\
         \midrule
                \rowcolor{light-yellow}
			ADMM-Net \cite{admm-net}
			& \tabincell{c}{Deep Unfolding\\(Supervised)}
               & ICCV 2019
			&\tabincell{c}{34.03\\0.919}
			&\tabincell{c}{33.57\\0.904}
			&\tabincell{c}{34.82\\0.933}
			&\tabincell{c}{39.46\\0.971}
			&\tabincell{c}{31.83\\0.924}
			&\tabincell{c}{32.47\\0.926}
			&\tabincell{c}{32.01\\0.898}
			&\tabincell{c}{30.49\\0.907}
			&\tabincell{c}{33.38\\0.917}
			&\tabincell{c}{30.55\\0.899}
			&\tabincell{c}{33.26\\0.920}
           \\
                \midrule
                \rowcolor{light-yellow}
			GAP-Net \cite{gapnet}
			& \tabincell{c}{Deep Unfolding\\(Supervised)}
               & IJCV 2023
			&\tabincell{c}{33.63\\0.913}
			&\tabincell{c}{33.19\\0.902}
			&\tabincell{c}{33.96\\0.931}
			&\tabincell{c}{39.14\\0.971}
			&\tabincell{c}{31.44\\0.921}
			&\tabincell{c}{32.29\\0.927}
			&\tabincell{c}{31.79\\0.903}
			&\tabincell{c}{30.25\\0.907}
			&\tabincell{c}{33.06\\0.916}
			&\tabincell{c}{30.14\\0.898}
			&\tabincell{c}{32.89\\0.919}
			\\
                \midrule
                \rowcolor{lightgreen}
			PnP-CASSI \cite{pnpcassi}
			& \tabincell{c}{PnP \\ (Zero-Shot)}
               & PR 2021
			&\tabincell{c}{29.09\\0.799}
			&\tabincell{c}{28.05\\0.708}
			&\tabincell{c}{30.15\\0.850}
			&\tabincell{c}{39.17\\0.939}
			&\tabincell{c}{27.45\\0.798}
			&\tabincell{c}{26.16\\0.752}
			&\tabincell{c}{26.92\\0.736}
			&\tabincell{c}{24.92\\0.710}
			&\tabincell{c}{27.99\\0.752}
			&\tabincell{c}{25.58\\0.664}
			&\tabincell{c}{28.55\\0.771}
			\\
			\midrule
                \rowcolor{lightgreen}
			DIP-HSI \cite{self}
			& \tabincell{c}{PnP \\ (Zero-Shot)}
               & ICCV 2021
			&\tabincell{c}{31.32\\0.855}
			&\tabincell{c}{25.89\\0.699}
			&\tabincell{c}{29.91\\0.839}
			&\tabincell{c}{38.69\\0.926}
			&\tabincell{c}{27.45\\0.796}
            &\tabincell{c}{29.53\\0.824}
			&\tabincell{c}{27.46\\0.700}
            &\tabincell{c}{27.69\\0.802}
			&\tabincell{c}{33.46\\0.863}
			&\tabincell{c}{26.10\\0.733}
			&\tabincell{c}{29.75\\0.803}
			\\
			\midrule
                \rowcolor{lightgreen}
			HLRTF \cite{hlrtf}
                &\tabincell{c}{Tensor Network\\(Self-Supervised)}
                & CVPR 2022
			&\tabincell{c}{29.04\\0.794}
			&\tabincell{c}{26.24\\0.679}
			&\tabincell{c}{27.91\\0.862}
			&\tabincell{c}{38.93\\0.962}
			&\tabincell{c}{25.33\\0.760}
			&\tabincell{c}{24.80\\0.735}
			&\tabincell{c}{25.54\\0.766}
			&\tabincell{c}{23.78\\0.733}
			&\tabincell{c}{30.04\\0.848}
			&\tabincell{c}{24.91\\0.660}
			&\tabincell{c}{27.65\\0.780}
                \\
                \midrule
			\rowcolor{lightgreen}
			\bf DiffSCI
			& \tabincell{c}{PnP-Diffusion\\(Zero-Shot)}
               & Ours
			&\tabincell{c}{34.96\\0.907}
			&\tabincell{c}{34.60\\0.905}
			&\tabincell{c}{\textbf{39.83}\\0.949}
			&\tabincell{c}{\underline{42.65}\\0.951}
			&\tabincell{c}{\textbf{35.21}\\0.946}
			&\tabincell{c}{33.12\\0.917}
			&\tabincell{c}{\textbf{36.29}\\ \underline{0.944}}
			&\tabincell{c}{30.42\\0.887}
			&\tabincell{c}{\textbf{37.27}\\0.931}
			&\tabincell{c}{28.49\\0.821}
			&\tabincell{c}{35.28\\0.916}
			\\
 
			\toprule[0.15em]
		\end{tabular}
	}
	\vspace{-6mm}
	\label{tab:simu}
\end{table*}

\vspace{-1mm}
\section{Experiments}
\label{sec:experiment}

\vspace{-2mm}
\subsection{Experiment Setup}
\vspace{-2mm}
Similar to most existing methods~\cite{tsa_net,hdnet,gsm,dauhst}, we select 10 scenes with spatial size 256$\times$256 and 28 bands from KAIST~\cite{kaist} as simulation dataset. Meanwhile, we select 5 MSIs with spatial size 660$\times$660 and 28 bands, captured by the CASSI system for real dataset~\cite{tsa_net}, then we crop data blocks of size 256$\times$256 for testing. The pre-trained diffusion model uses a model trained by~\cite{zhu2023denoising}.   

\noindent\textbf{Parameter Setting.}
Through all our experiments, we use the same linear noise schedule $\{\beta_t\}$, and DDIM sampling. The shift step is set to 2. And in the wavelength matching method, we choose $21_{th}$ and $28_{th}$ bands to form a three-channel image. Meanwhile, we set the reverse initial time step to 600 and set the sampling steps to 20, 100, 200, 300 and 500 respectively for testing. After experiments, we find setting $\lambda = 15$, $\eta = 1$, $\zeta = 1$ in DDIM process and $sc = 1$ in data proximal subproblem can achieve the best results.

\noindent\textbf{Comparisons with SOTA Methods.}
In this section, we test the performance of our proposed DiffSCI method on the simulation dataset. We compare the results of our DiffSCI method with 15 SOTA methods including three model-based methods (TwIST~\cite{twist}, GAP-TV~\cite{gap_tv}, DESCI~\cite{desci}), six E2E methods ($\lambda$-Net~\cite{lambda}, TSA-NET~\cite{tsa_net}, HDNET~\cite{hdnet}, MST-L~\cite{mst}, MST++~\cite{mst_pp}, CST-L-PLUS~\cite{cst}), three deep unfolding methods (DGSMP~\cite{gsm}, GAP-NET~\cite{gapnet}, ADMM-NET~\cite{admm-net}), two PnP methods (PnP-CASSI~\cite{pnpcassi}, DIP-MSI~\cite{self}) and one tensor network method (HLRTF~\cite{hlrtf}) on 10 simulation scenes with the same settings. From Table~\ref{tab:simu}, it can be observed that our unsupervised method has a significant improvement compared to other unsupervised methods. The gap between its performance on PSNR and current supervised SOTA methods such as MST-L~\cite{mst} and MST++~\cite{mst_pp} is also narrowing. Moreover, we do not need to retrain a model on MSIs. Therefore, the proposed DiffSCI achieves a balance between flexibility and performance.

\begin{figure*}
	\vspace{-10mm}
	\centering
	\renewcommand{\h}{0.105}
	\renewcommand{\wa}{0.12}
	\newcommand{\wb}{0.16}
	\renewcommand{\g}{-0.7mm}
	\renewcommand{\tabcolsep}{1.8pt}
	\renewcommand{\arraystretch}{1}
        \resizebox{1\linewidth}{!} {
        	\includegraphics[width=0.47\linewidth, angle =270]{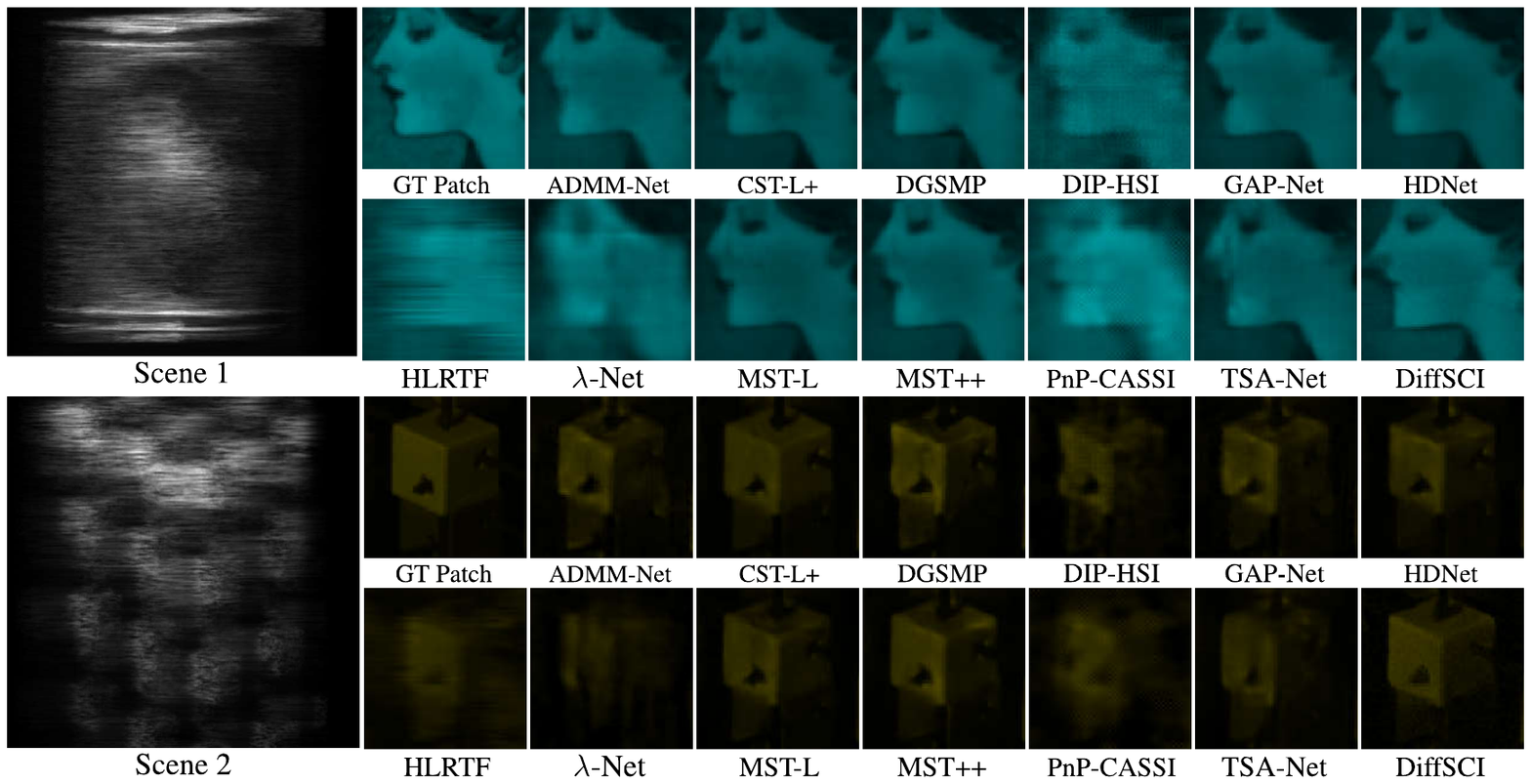} }
	\vspace{-20mm}
	\caption{Visual comparison on KAIST dataset. \textbf{Top} is $Scene$ 1 at wavelength 487.0nm. \textbf{Bottom} is $Scene$ 2 at wavelength 575.5nm} %
    \vspace{-6mm}
	\label{fig_kaist_s1}
\end{figure*}

\begin{figure}
\centering
\includegraphics[width=0.47\linewidth]{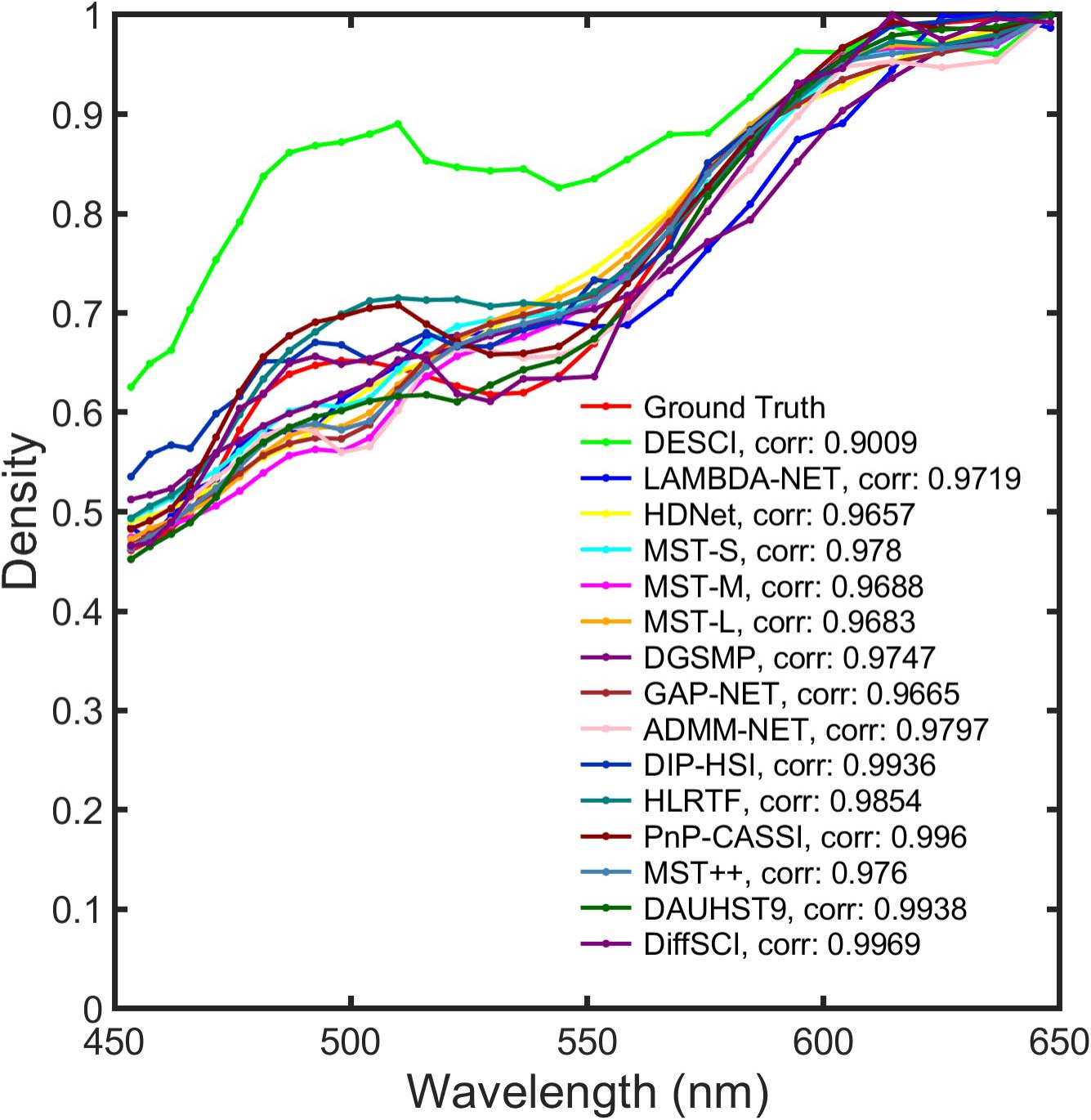}
\includegraphics[width=0.47\linewidth]{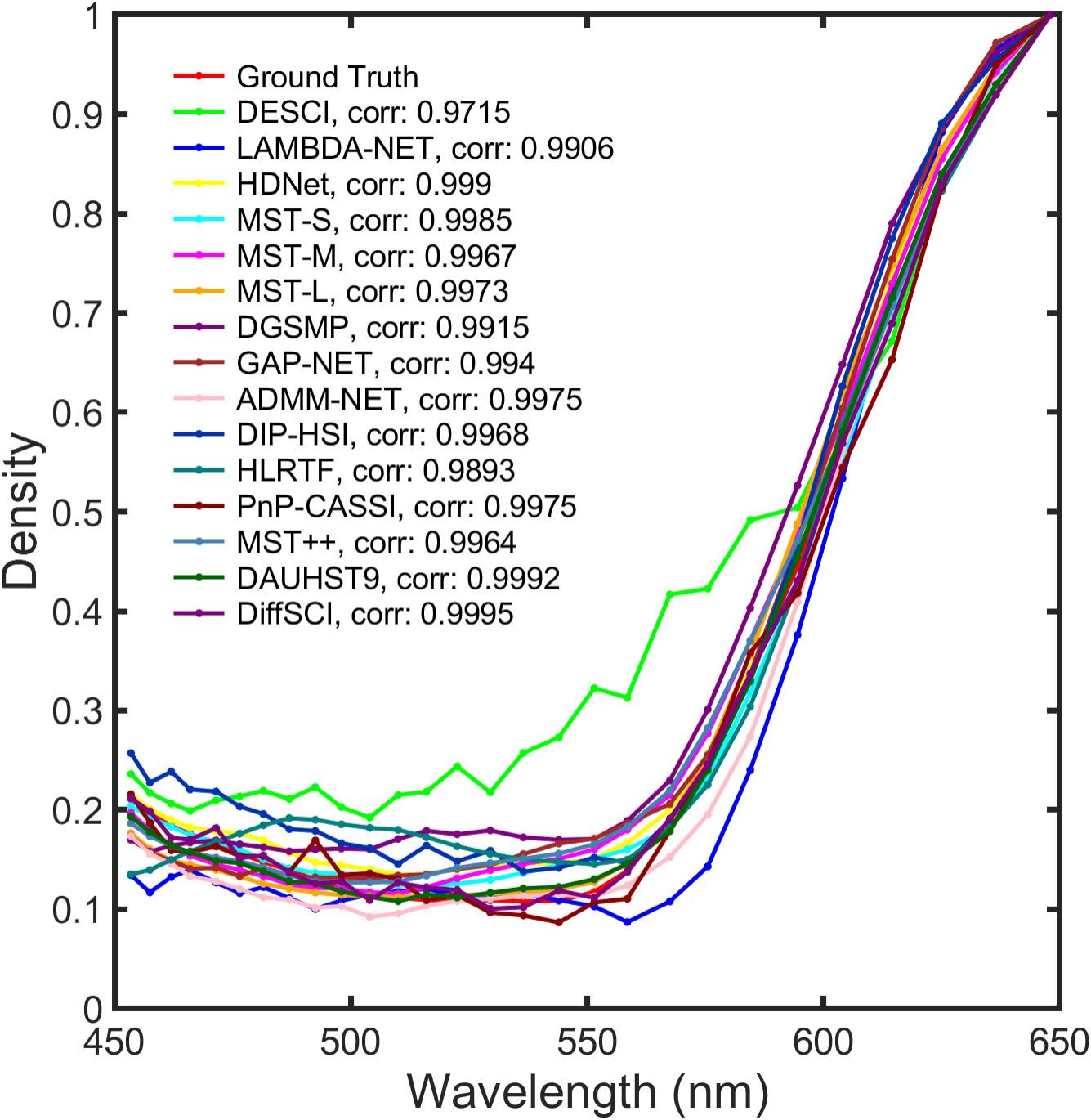}
\vspace{-3mm}
   \caption{Spectral Density Curves.}
\label{fig_GSA}
\vspace{-8mm}
\end{figure}

\vspace{-2mm}
\subsection{Qualitative Experiments}
\vspace{-2mm}
\textbf{Results on Simulation Dataset.}
Fig.~\ref{fig_kaist_s1} shows the display effects of MSI reconstruction between our DiffSCI method and other SOTA methods on the $8_{th}$ band of $Scene$ 1 (top) and $21_{th}$ band of $Scene$ 2 (bottom). From the enlarged part of the $Scene$ 1 image, we can see that our DiffSCI provides superior visual effects of detailed contents, cleaner textures, and fewer artifacts compared to other SOTA methods. Furthermore, to demonstrate the powerful generative capabilities of the diffusion model, we can observe the magnified section of $Scene$ 2. Our method makes the edges of the blocks sharper, the shapes and patterns closer to the GT, whereas previous methods either generate over-smoothed results thus losing the complexity of fine-grained structures or introduce artifacts. This suggests that the generative capabilities of diffusion can be effectively applied to reconstruct darker regions, thereby filling in the gaps in the current method. Fig.~\ref{fig_GSA} presents the density-wavelength spectral curves. The spectral curves from DiffSCI achieve the highest correlation with the reference curves, even exceeding the performance of the current leading method, DAUHST-9~\cite{dauhst}. This demonstrates the superiority of our proposed DiffSCI in terms of spectral-dimension consistency.

\begin{figure*}
	\vspace{-32mm}
	\centering
	\renewcommand{\h}{0.105}
	\renewcommand{\wa}{0.12}
	\newcommand{\wb}{0.16}
	\renewcommand{\g}{-0.7mm}
	\renewcommand{\tabcolsep}{1.8pt}
	\renewcommand{\arraystretch}{1}
        \resizebox{1\linewidth}{!} {\includegraphics[width=0.47\linewidth, angle =270]{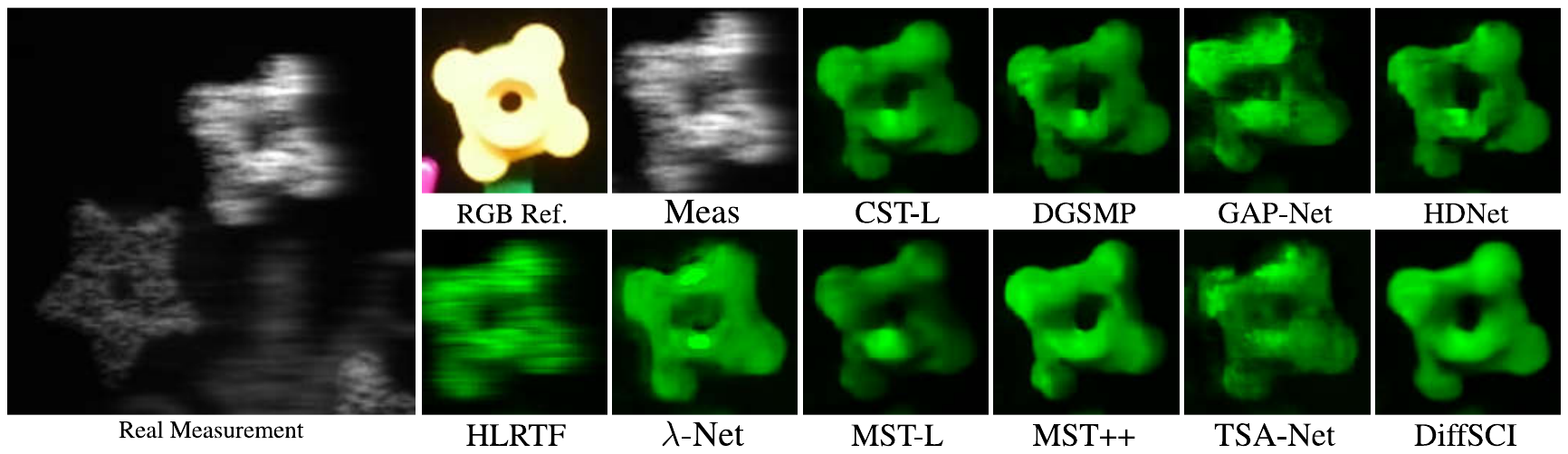}}
	\vspace{-39mm}
	\caption{Visual comparison of SCI reconstruction methods on $Scene$ 1 of real dataset at wavelength 536.6nm.} %
	\label{fig_real_s1}
    \vspace{-4mm}
\end{figure*}

\begin{figure}
	\vspace{-8mm}
	\centering
	\renewcommand{\h}{0.105}
	\renewcommand{\wa}{0.12}
	\newcommand{\wb}{0.16}
	\renewcommand{\g}{-0.7mm}
	\renewcommand{\tabcolsep}{1.8pt}
	\renewcommand{\arraystretch}{1}
   \resizebox{1\linewidth}{!} {\includegraphics[width=0.99\linewidth, angle =270]{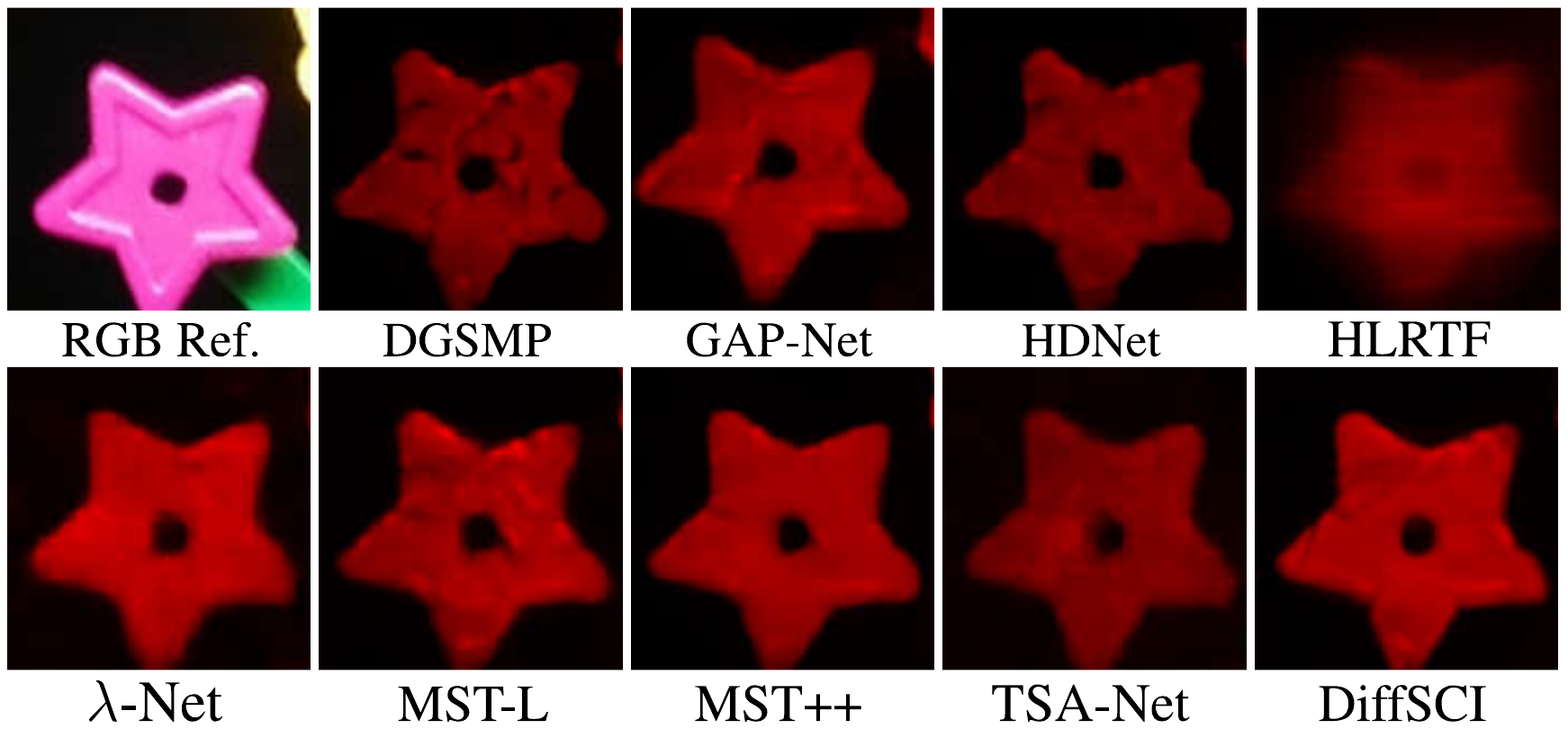}}
	\vspace{-14mm}
	\caption{Visual comparison on $Scene$ 1 of real dataset.} %
	\label{fig_real_s1_28}
    \vspace{-7mm}
\end{figure}

\noindent\textbf{Results on Real Dataset.}
We also test the reconstruction capability of DiffSCI on real dataset. Fig.~\ref{fig_real_s1} and Fig.~\ref{fig_real_s1_28} show the visual comparison between DiffSCI and other SOTA methods. It is evident that our reconstruction results are more detailed and have fewer artifacts. Compared to the blurred results reconstructed by other methods, our method DiffSCI demonstrates that the generative ability of the diffusion model can provide good robustness against noise, leading to enhanced results in MSI reconstruction. More experimental results are shown in the supplementary materials.

\vspace{-2mm}
\section{Ablation Study}
\label{sec:ablation}
\vspace{-2mm}
\textbf{Effects of Acceleration Algorithm.}
We propose a residual accumulation method aimed at achieving acceleration. Through experimentation, employing this accelerated algorithm showcases enhancements not only in convergence speed but also in performance, maintaining consistent parameters. Figure~\ref{fig:acc} demonstrates the impact of the acceleration algorithm on both PSNR and time, utilizing an identical number of sampling steps. Evidently, the accelerated algorithm yields an improvement of 5-6dB in average performance while expediting convergence.

\begin{figure}
  \centering
  \begin{overpic}[width = 0.48\textwidth]{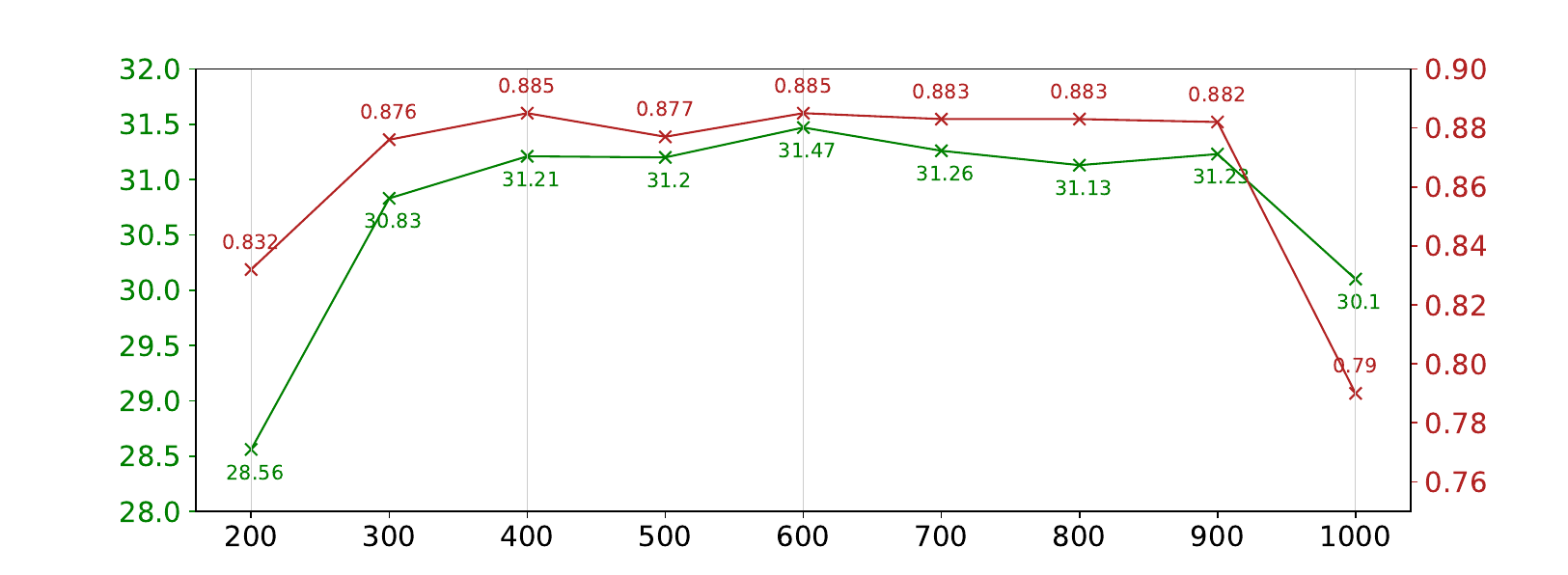}
  \put(40,-3){\color{black}{\scriptsize $t_{start}$ (steps) }}
  \put(-3,13){\rotatebox{90}{\scriptsize PSNR/db}}
  \put(100,15){\rotatebox{90}{\scriptsize SSIM}}
  \end{overpic}
  
  \vspace{10pt} 
  
  \begin{subfigure}[b]{0.1\textwidth}
    \centering
    \includegraphics[width=\textwidth]{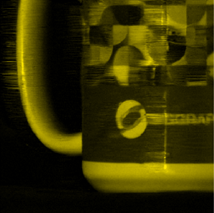}
    \caption{200}
    \label{fig:small_image1}
  \end{subfigure}
    \hspace{1mm}
  \begin{subfigure}[b]{0.1\textwidth}
    \centering
    \includegraphics[width=\textwidth]{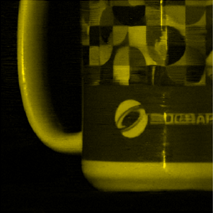}
    \caption{400}
    \label{fig:small_image2}
  \end{subfigure}
     \hspace{1mm}
  \begin{subfigure}[b]{0.1\textwidth}
    \centering
    \includegraphics[width=\textwidth]{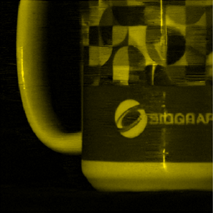}
    \caption{600}
    \label{fig:small_image3}
  \end{subfigure}
     \hspace{1mm}
  \begin{subfigure}[b]{0.1\textwidth}
    \centering
    \includegraphics[width=\textwidth]{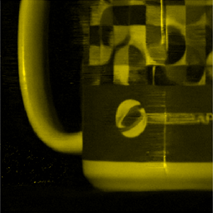}
    \caption{1000}
    \label{fig:small_image4}
  \end{subfigure}
  \label{fig:four_small_images}
  \vspace{-4mm}
  \caption{Effect of $t_{start}$ on $Scene$ 5 of KAIST.}
  \label{fig:start}
  \vspace{-3mm}
\end{figure}

\noindent\textbf{Effects of $t_{start}$.}
Our DiffSCI can perform the reverse process from partially noisy images instead of starting the recovery from pure Gaussian noise. To demonstrate the impact of $t_{start}$ on performance briefly, we show how PSNR changes in Fig.~\ref{fig:start}. We select sampling steps with 100 for all experiments and find that our method achieves the best results in terms of PSNR and SSIM at $t_{start} = 600$.

\noindent\textbf{Effects of Sampling Steps.}
To study the impact of the number of sampling steps on the reconstruction quality assessment parameters PSNR and SSIM, and thus balance the sampling speed with the recovery quality, we conduct experiments with different numbers of sampling steps. As shown in Fig.~\ref{fig:acc}, we set sampling steps $T \in [20,100,200,300,500]$.

\begin{figure}
\centering
\begin{overpic}[height = 0.47\linewidth,width=0.46\linewidth]{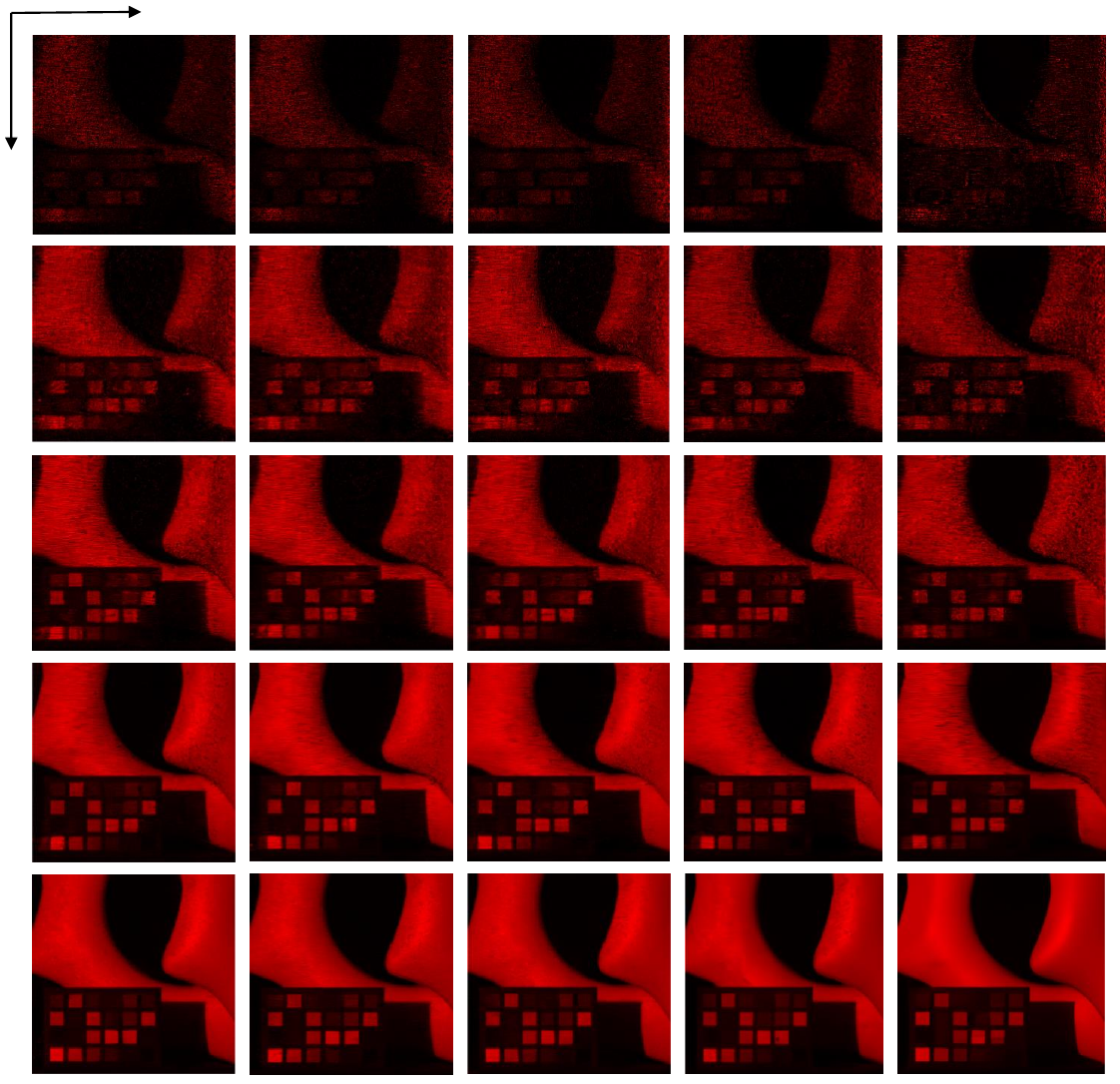}
\put(1,100){\color{black}{\scriptsize $\lambda$}}
\put(-4,95){\color{black}{\scriptsize $\zeta$}}
\put(9,100){\color{black}{\scriptsize 0.1}}
\put(25,100){\color{black}{\scriptsize 1.0}}
\put(45,100){\color{black}{\scriptsize 10}}
\put(62,100){\color{black}{\scriptsize 100}}
\put(80,100){\color{black}{\scriptsize 1000}}
\put(-9,10){\color{black}{\scriptsize 1.0}}
\put(-9,28){\color{black}{\scriptsize 0.8}}
\put(-9,48){\color{black}{\scriptsize 0.5}}
\put(-9,68){\color{black}{\scriptsize 0.2}}
\put(-9,86){\color{black}{\scriptsize 0.0}}
\end{overpic}
\hspace{2mm}
\begin{overpic}[height = 0.47\linewidth,width=0.48\linewidth]{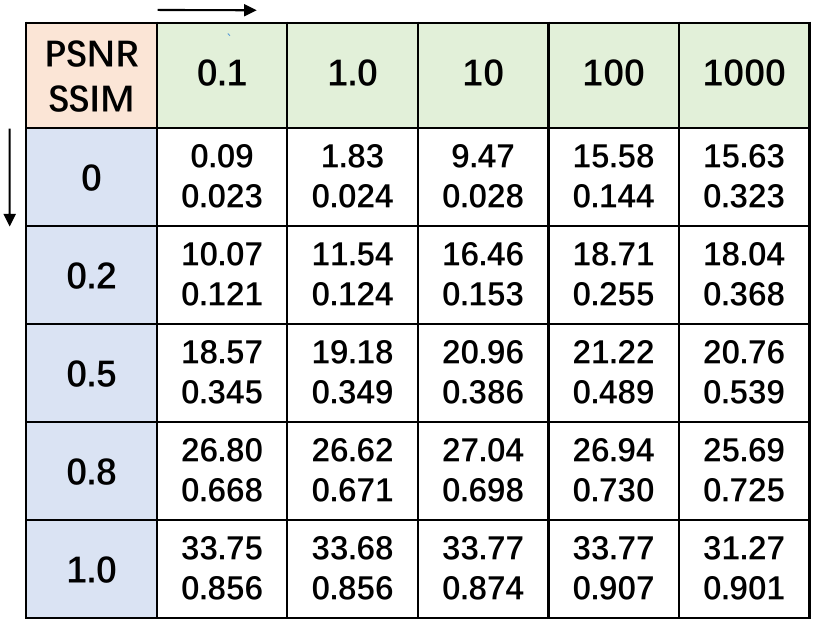}
\put(21,98){\color{black}{\scriptsize $\lambda$}}
\put(-4,70){\color{black}{\scriptsize $\zeta$}}
\end{overpic}
\vspace{-3mm}
        \caption{ Visual comparison (left) and PSNR comparison (right) of the effect of hyperparameters $\zeta$ and $\lambda$ on $Scene$ 3 of KAIST.}
    \label{fig:para}
    \vspace{-3mm}
\end{figure}

\begin{figure}
\begin{subfigure}[b]{0.11\textwidth}
    \centering
    \includegraphics[width=\textwidth]{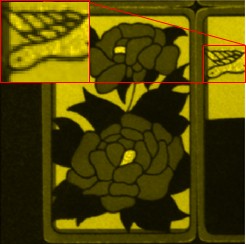}
    \caption{GT}
    \label{fig:sc_GT}
  \end{subfigure}
    \hspace{0.5mm}
  \begin{subfigure}[b]{0.11\textwidth}
    \centering
    \includegraphics[width=\textwidth]{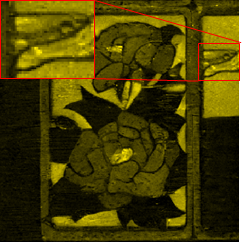}
    \caption{0.5}
    \label{fig:sc_0.5}
  \end{subfigure}
     \hspace{0.5mm}
  \begin{subfigure}[b]{0.11\textwidth}
    \centering
    \includegraphics[width=\textwidth]{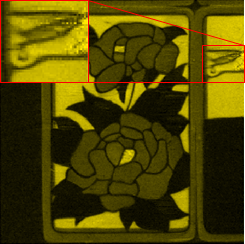}
    \caption{1.0}
    \label{fig:sc_1.0}
  \end{subfigure}
     \hspace{0.5mm}
  \begin{subfigure}[b]{0.11\textwidth}
    \centering
    \includegraphics[width=\textwidth]{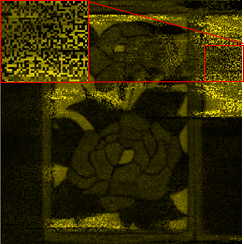}
    \caption{2.0}
    \label{fig:sc_2.0}
  \end{subfigure}
  \vspace{-8mm}
        \caption{Effect of $sc$ (0.5, 1.0, 2.0) on $Scene$ 7 of KAIST.}
        \vspace{-4mm}
    \label{fig:guidance}
\end{figure}

\noindent\textbf{Effects of $\lambda $, $\zeta$ and $sc$.} DiffSCI has three hyperparameters $\lambda$, $\zeta$ and $sc$, which manage the strength of the condition guidance, the level of noise added at each timestep and the update step size in data proximity subproblems. As shown in the left figure of Fig.~\ref{fig:para}, when $\zeta$ approaches 1, we get the best reconstruction quality. Meanwhile, when $\lambda < 1$, which means the condition guidance is strong enough, the reconstructed MSI amplifies noise, and when $\lambda = 1000$ the reconstructed MSI becomes unconditional. Shown in the right figure, we find values of $\lambda$ that are too large or small will impact the PSNR. Meanwhile, Fig.~\ref{fig:guidance} demonstrates a close relationship between $sc$ and the quality of the reconstruction. Good reconstruction results can be achieved when $sc$ achieves 1.
Too small or too large step sizes would lead to reconstruction distortion.

\vspace{-3mm}
\section{Conclusion}
\label{sec:conclusion}
\vspace{-2mm}
In this paper, we are the first to integrate diffusion model with Plug-and-Play algorithm, applying the generative capabilities of the diffusion model to MSI reconstruction, which compensated for the shortcomings of current methods. Specifically, by utilizing the wavelength matching method and HQS method, we successfully applied the HQS-based diffusion model, which was pre-trained on RGB images, as a denoising prior in MSI reconstruction. Meanwhile, we introduced acceleration algorithms when solving the data subproblem. Experimental results on both simulated and real datasets highlighted the superior adaptability, efficiency, and applicability of DiffSCI compared to SOTA methods.




{\small
\bibliographystyle{ieeenat_fullname}
\bibliography{11_references}
}

\ifarxiv \clearpage \appendix \input{12_appendix} \fi
\end{document}